\documentclass[runningheads]{llncs}
\usepackage[T1]{fontenc}
\usepackage{graphicx}
\usepackage[hidelinks]{hyperref}
\usepackage{color}

\begin{document}
\title{Do Echo Top Heights Improve Deep Learning Rainfall Nowcasts? A Case Study in the Netherlands}
\titlerunning{Do Echo Top Heights Improve Deep Learning Nowcasts?}
\author{Peter Pavlík\inst{1, 2} \orcidID{0000-0002-7468-5503}
\and Marc Schleiss\inst{3} \orcidID{0000-0002-4667-5500}
\and Anna Bou Ezzeddine\inst{1} \orcidID{0000-0002-3341-6059}
\and Viera Rozinajová\inst{1} \orcidID{0000-0003-1302-6261}}
\authorrunning{P. Pavlík, M. Schleiss et al.}
%
\institute{Kempelen Institute of Intelligent Technologies, Bratislava, Slovakia \and
Faculty of information Technology, Brno University of Technology, Czech Republic \and
Department of Geoscience and Remote Sensing, Delft University of Technology, Netherlands}
\maketitle              
\begin{abstract}
Precipitation nowcasting -- the short-term prediction of rainfall using recent radar observations -- is critical for weather-sensitive sectors such as transportation, agriculture, and disaster mitigation. While recent deep learning models have shown promise in improving nowcasting skill, most approaches rely solely on 2D radar reflectivity fields, discarding valuable vertical information available in the full 3D radar volume. In this work, we explore the use of Echo Top Height (ETH), a 2D projection indicating the maximum altitude of radar reflectivity above a given threshold, as an auxiliary input variable for deep learning-based nowcasting. We examine the relationship between ETH and radar reflectivity, confirming its relevance for predicting rainfall intensity. We implement a single-pass 3D U-Net that processes both the radar reflectivity and ETH as separate input channels. While our models are able to leverage ETH to improve skill at low rain-rate thresholds, results are inconsistent at higher intensities and the models with ETH systematically underestimate precipitation intensity. Three case studies are used to illustrate how ETH can help in some cases, but also confuse the models and increase the error variance. Nonetheless, the study serves as a foundation for critically assessing the potential contribution of additional variables to nowcasting performance.

\keywords{Precipitation Nowcasting  \and Deep Learning \and Weather Radar.}
\end{abstract}

\section{Introduction}

Nowcasting is defined by the World Meteorological Agency as forecasting with local detail, by any method, over a period from the present to six hours ahead, including a detailed description of the present weather~\cite{WMO17}. In this paper, we focus on the task of nowcasting precipitation amounts in the near future for the purpose of generating alerts on extreme weather events and preventing damage to infrastructure due to flooding. Previous work on precipitation nowcasting has shown that even at small lead times of less than an hour, accurate precipitation prediction can be challenging. In fact, precipitation is one of the most difficult weather variables to accurately forecast~\cite{sun2014use}.

Traditional nowcasting techniques rely on radar reflectivity data and optical flow–based extrapolation methods to estimate the future position of precipitation fields. While computationally efficient, these methods struggle with predicting the initiation, growth, or dissipation of rain cells, particularly during convective storms. In recent years, deep learning has emerged as a promising approach to address these limitations by directly learning the complex spatiotemporal dynamics of precipitation from sequences of radar observations.

This paper investigates the potential of incorporating additional vertical atmospheric structure information — specifically, echo top height (ETH) — into radar-based deep learning nowcasting models. ETH is a variable derived from volumetric radar scans representing the top altitude of detected precipitation that may offer insight into the vertical development of rain cells, potentially enabling earlier detection of intensifying convection. We begin by exploring the relationship between ETH and radar reflectivity (dBZ) and establish that ETH contains complementary, though not redundant, information.

To assess the utility of ETH, we implemented several variants of U-Net architectures that incorporate the ETH variable in different ways. Among the tested models, we identified a 3D U-Net architecture as being the most suitable model for jointly learning from radar reflectivity and ETH data. This particular architecture allows to represent radar reflectivity and ETH as separate data channels and provides a natural way to incorporate information about the vertical structure of precipitation.

\section{Precipitation Nowcasting -- An Overview}

The short-term prediction of precipitation intensity and location -- called precipitation nowcasting -- is paramount for many critical applications. It is crucial for taking effective measures aimed towards disaster prevention and mitigation and making decisions in agriculture, transportation and other weather dependent domains. It is increasingly vital when facing the climate crisis which increases the intensity and variability of precipitation~\cite{Zhang2024}.

The short lead times in nowcasting strongly limit the type of data and methods that can be leveraged. Mid-to-long-term forecasts are operationally calculated by simulating the future state of the atmosphere using numerical weather prediction (NWP) models. However, these models are computationally expensive to run and cannot be used to quickly assimilate new data due to their large spin-up times~\cite{liu2013study}. By the time the simulation outputs are available, we may have already overshot the short-term nowcast target. Consequently, nowcasting methods mostly rely on simpler, data-driven methods that approximate the physical processes in the atmosphere but are fast to run and can be updated very often (e.g., every 5 minutes).

For precipitation nowcasting, the most common data sources are ground radar reflectivity observations. Weather radar can be used to monitor precipitation with high spatial detail and a high update rates of just a few minutes~\cite{WMO17}, making them the best fit for the task.

Since radar data are commonly mapped to a Cartesian 2-dimensional grid, many methods from the field of computer vision are applicable. For example, autoregressive models can be used to estimate the spatial correlation structure of past and current precipitation fields, from which the next images can be predicted, similar to the next frame video prediction problem. Typically, this also involves a motion estimation and extrapolation step, during which precipitating cells are advected along the principal direction of motion. The motion field itself can be estimated from an optical flow or cross-correlation algorithm, such as in TITAN~\cite{dixon1993titan}, COTREC~\cite{li1995nowcasting}, STEPS~\cite{bowler2006steps} and others. A major shortcoming of these extrapolation-based nowcasting methods is that the precipitation field is assumed to only change very slowly over time as it moves over different areas, which may not be true in very dynamic weather conditions such as convective rain. While modern nowcasting approaches try to tackle this in various ways, there are fundamental limits to what can be predicted based on radar data alone, and most dynamic processes such as the growth, intensification and decay of rain cells, or more generally, any change in rain cell size or motion, currently cannot be reliably predicted~\cite{gmd-12-4185-2019}.

In the past years, deep-learning approaches have started to gain attention by attempting to mitigate these main limitations and learn to predict the complex space-time dynamics of precipitation fields. Deep learning models can learn complex patterns while still satisfying the requirement of producing an output quickly, offering the potential to overcome many of the limitations inherent in traditional methodologies. However, these approaches come with their own set of challenges.

\subsection{Deep Learning for Precipitation Nowcasting}

Within the realm of deep learning for precipitation nowcasting, the problem is often framed as a spatiotemporal sequence prediction task, where the goal is to forecast future radar precipitation maps based on a sequence of past observations. This perspective allows researchers to adapt and apply various deep learning architectures that have proven successful in video prediction and other sequence modeling domains.

The first notable deep learning model used for precipitation nowcasting was the ConvLSTM model presented in~\cite{shi2015convolutional}, which integrates convolutional layers into the recurrent LSTM architecture, enabling the model to effectively capture both spatial and temporal correlations within successive radar precipitation maps. A limitation of this approach is the complex motion of precipitation over time that results in location variance of precipitation patterns in the successive observations (the corresponding precipitation object moves to a different part of the successive image), which the location-invariant convolution filters are inadequate for. The Traj-GRU model from~\cite{shi2017deep} builds upon the ConvLSTM and solves the limitation by allowing the model to learn location-variant recurrent connections dynamically. It employs a subnetwork that generates flow fields determining the sampling locations for the hidden states dynamically and allows TrajGRU to capture complex spatiotemporal correlations more effectively than ConvLSTM.

Despite the sequential character of the data, some researchers decided to abandon the recurrent neural network architecture for simpler convolutional-only models. A widely used architecture is U-Net -- an encoder–decoder structure with skip connections. Its first notable use for precipitation nowcasting was RainNet in~\cite{ayzel2020rainnet}, treating nowcasting as an image-to-image translation problem. RainNet is trained to predict only a single next observation. However, longer forecasts can also be obtained by recursively applying the same model and using previous predictions as inputs. Previous research has shown that this can lead to cumulative smoothing effects, and a general loss of detail in predicted precipitation patterns over time and intensity degradation~\cite{ayzel2020rainnet}. However, these limitations can be mitigated by dropping the iterative approach and predicting all the required lead times at once.

The blurriness issue is not limited to RainNet. It affects the vast majority of machine learning-based precipitation nowcasting models that generate a single deterministic prediction. The increasingly blurry output with increasing lead time is a major issue when trying to forecast heavy rain. The root cause is the usage of gridpoint-based error metrics as loss functions to minimize, which results in the so-called `double penalty problem'. A forecast of a precipitation feature that is correct in terms of intensity, size, and timing, but incorrect in its location, causes a very large error~\cite{Keil2009}. This heavily interferes with the main goal of the nowcasting models -- prediction of heavy precipitation events associated with significant societal impacts -- as the highest intensity values in the data degrade over time.

The blurriness problem, inherent to deterministic models predicting the evolution of highly chaotic and uncertain precipitation fields, led to the introduction of adversarial learning approaches (GANs) such as in the DGMR model~\cite{ravuri2021skilful}. The latter aim to create visually plausible outputs without blurring over time by incorporating one or several discriminators. Another useful addition is to incorporate the underlying physical laws governing the behavior of rain through so-called ``physics-informed machine learning'' like in~\cite{zhang2023skilful}. Such models are trained to predict dynamic extrapolation motion fields similar to the TrajGRU, while making sure that the predictions satisfy some basic physical or mathematical property such as the continuity equation. This ensures that the outputs are not just visually plausible, but also physically consistent with the laws of nature.

However, just like deterministic models can produce blurred predictions, GANs are prone to many issues, such as training instability and the generation of artifacts. Some models like CasCast~\cite{gong2024cascast} produce a deterministic prediction that then serves as an input to a generative diffusion model, thereby trying to combine the two approaches by capitalizing on their respective strengths while minimizing their downsides.  

While deep learning methods have pushed the limits of what we thought was possible in operational rainfall nowcasting~\cite{zhang2023skilful,espeholt2022deep}, many fundamental issues remain. Modeling the complex evolution of precipitation systems across different spatial and temporal scales and achieving accurate forecasts for extreme precipitation events remain key areas of ongoing research. To address these limitations, there is a growing interest in incorporating additional physical constraints and data sources into deep learning models, to help them better predict the non-linear dynamics of rain cells.

\paragraph{The Potential of Additional Data Variables.}

To obtain a useful forecast at higher lead times, additional attributes about the weather not captured by the radar reflectivity maps are required. Key atmospheric variables such as air pressure, wind and temperature which heavily affect future weather evolution should be considered. Presumably cloud type, vertical profiles of temperature and pressure would help as well, but it's not clear how much.

Additionally, there are many factors that affect the utility of additional variables. First, to include data from other atmospheric sensors, we might need to perform computationally expensive data assimilation. This increases the total computation time considerably. For the task of nowcasting, we need to carefully balance the benefits that the additional data sources can provide with the cost of their inclusion at operational runtime (gathering the various observations and their spatio-temporal mapping). An equally important aspect to be considered is the fact that surface data on their own are insufficient to characterize and predict the dynamics of clouds and precipitation. Unfortunately, the availability of vertical profiles and 3D atmospheric observations is limited, especially in real-time.

Another issue is the fact that the relationship between the primary and secondary variables in the model might change over time, depending on other, hidden factors or variables. Measurement errors and sampling uncertainty might also be an issue. In the worst case scenario, additional input channels may actually confuse the models and cause the predictions to become less reliable. The prediction error might be slightly lower but the fluctuations of the prediction errors over time and across events might be larger than before, which is undesirable.

Another crucial point deserves attention. Most machine learning approaches to nowcasting implicitly assume a fixed, stationary relationship between inputs and outputs. However, in the context of precipitation nowcasting - where chaotic, highly dynamic atmospheric processes are at play - this assumption often breaks down. Key information necessary to accurately predict future developments is often missing or unobservable. As a result, while a model might successfully learn the average mapping between inputs and outputs over a training dataset, this has limited practical value if it cannot adapt to individual, especially extreme, cases. These rare events often fall outside the learned average behavior, yet they are precisely the ones that need to be forecasted accurately.

\subsection*{Section Summary}

Precipitation nowcasting predicts rain in the near future and is critical for timely decisions in weather-sensitive sectors. Traditional NWP models are too slow for short-term use, so fast methods working with radar observations are preferred. Deep learning models can be used for nowcasting, but they struggle with issues like blurriness, instability, and performance issues on rare extreme events. Incorporating extra atmospheric data may help but may also confuse the models.

\section{Radar Reflectivity and Echo Top Height}

Most existing radar-based nowcasting approaches rely solely on 2D radar reflectivity representations, such as constant altitude plan position indicators (CAPPI) or vertically integrated maxima (CMAX). While effective, these projections discard much of the vertical information captured by weather radars, which operate by scanning at multiple elevation angles and thus provide a full 3D view of the precipitation structure surrounding the radar station. Recent work~\cite{pavlik2022radar} has demonstrated that incorporating this full 3D volumetric data can improve nowcasting accuracy.

\begin{figure}
  \centering
  \includegraphics[width=\linewidth]{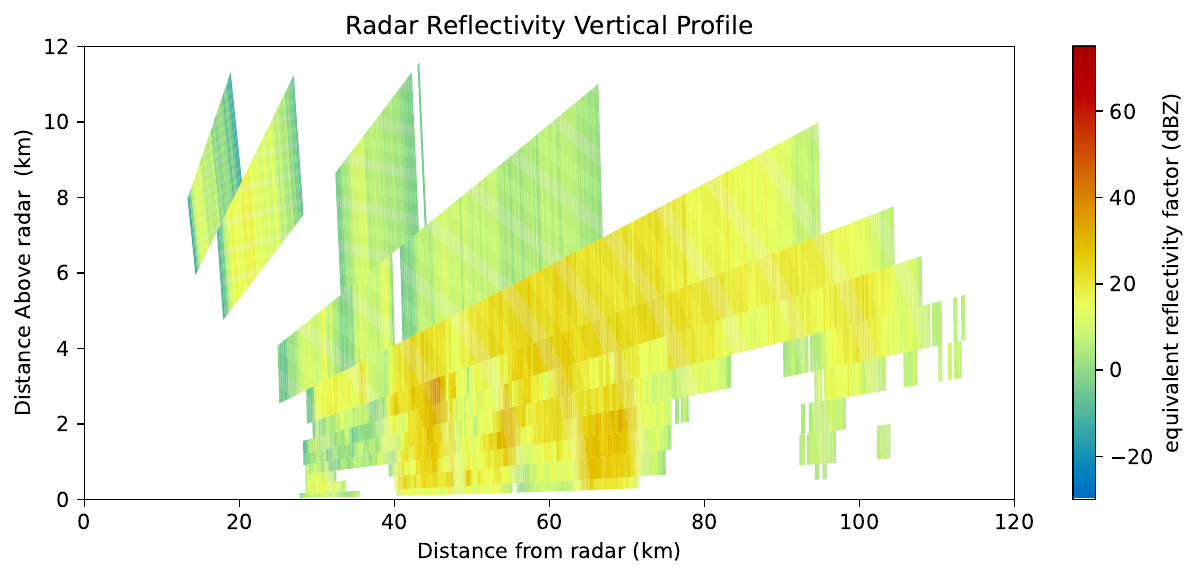}
  \caption{This Example shows a single vertical slice of the full 3-dimensional radar observation. The radar station is located the the bottom left corner. Each ray represents data captured by a different radar sweep.}
  \label{fig:reflectivity_vert}
\end{figure}

However, working with a full 3D volume can be very computationally expensive. Also, if we consider the time steps as a separate data dimension, we effectively arrive at 4D data volumes that the model needs to process and many popular machine learning frameworks do not even support 4D convolution operations natively. As a result, there is a need for compact yet informative representations of vertical storm structure that can be more easily integrated into existing 2D-based models. This is where Echo Top Height comes in.

\begin{figure}[!h]
  \centering
  \includegraphics[width=\linewidth]{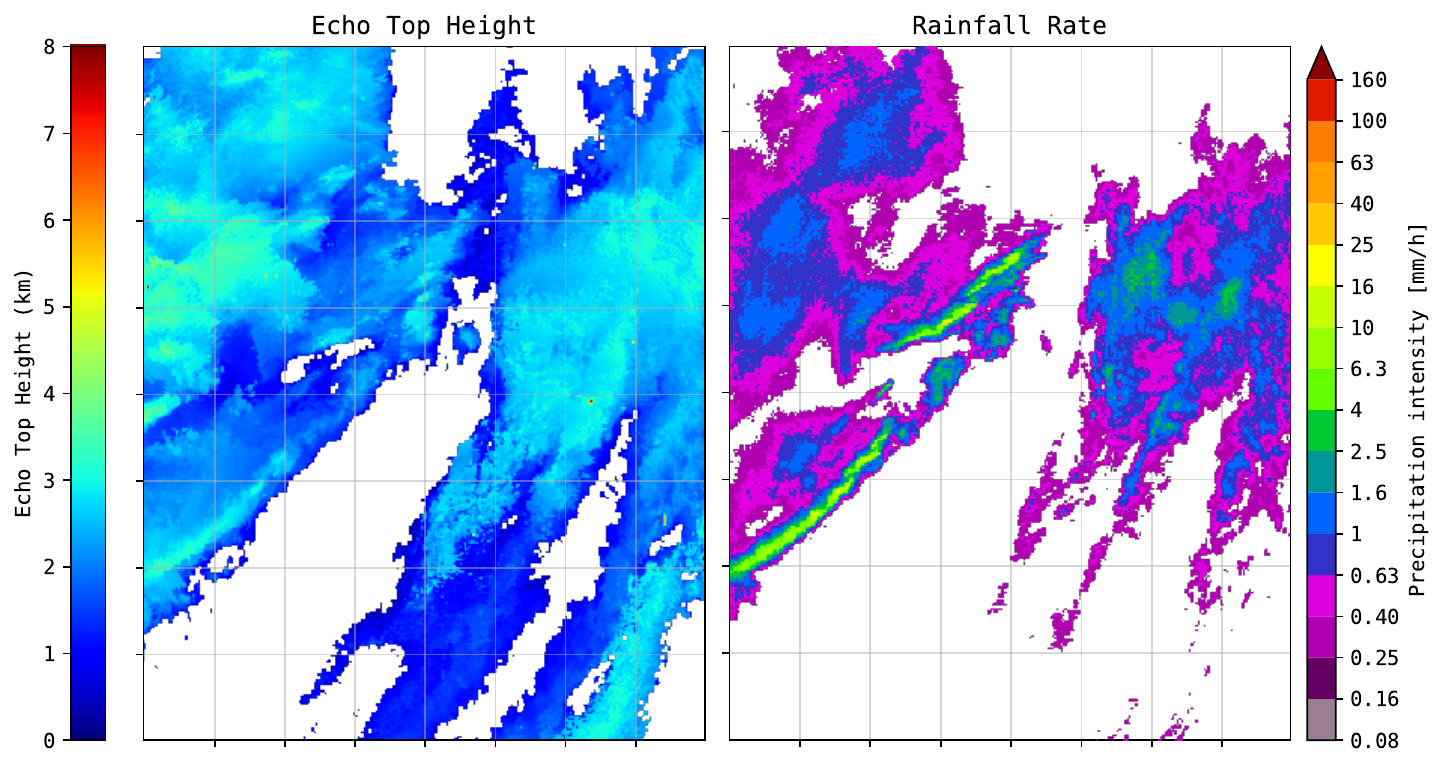}
  \caption{Side-by-side comparison of a single observation from the dataset (February 20th, 2022), showing the spatial distribution of echo top height (left) and corresponding rainfall rate (right). This illustrates the relationship between storm structure aloft and precipitation intensity at 1500 meters above ground. In this case, a reflectivity threshold of 7 dBZ was used to calculate echo top height.}
  \label{fig:ETH}
\end{figure}

Echo top height (ETH) refers to the maximum altitude at which a weather radar detects a reflectivity value above a certain threshold. This measurement serves as an indicator of the storm's vertical extent and intensity, and can be used to provide insight into a storm's development and severity. Various reflectivity thresholds such as 7 or 18 dBZ are used depending on the specific application and radar system. See Figure~\ref{fig:ETH} for an example observation.

\begin{figure}[!h]
    \centering
    \includegraphics[width=0.49\linewidth]{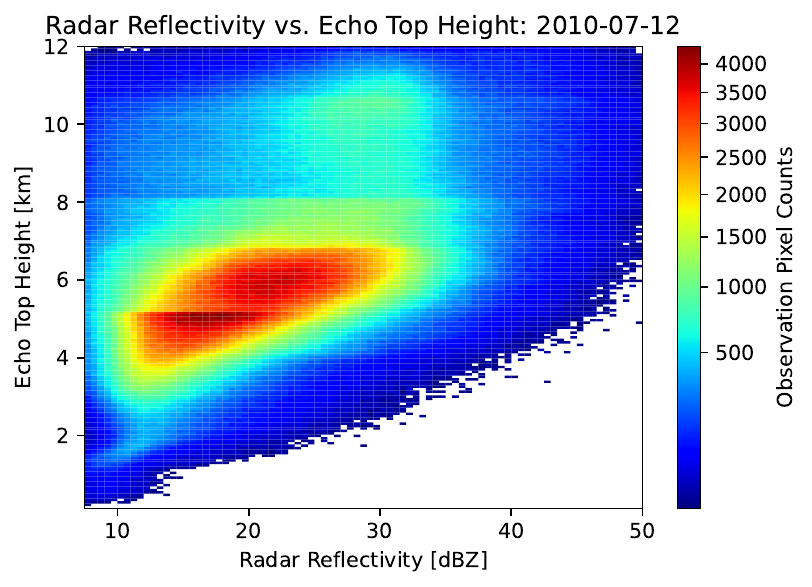}
    \includegraphics[width=0.49\linewidth]{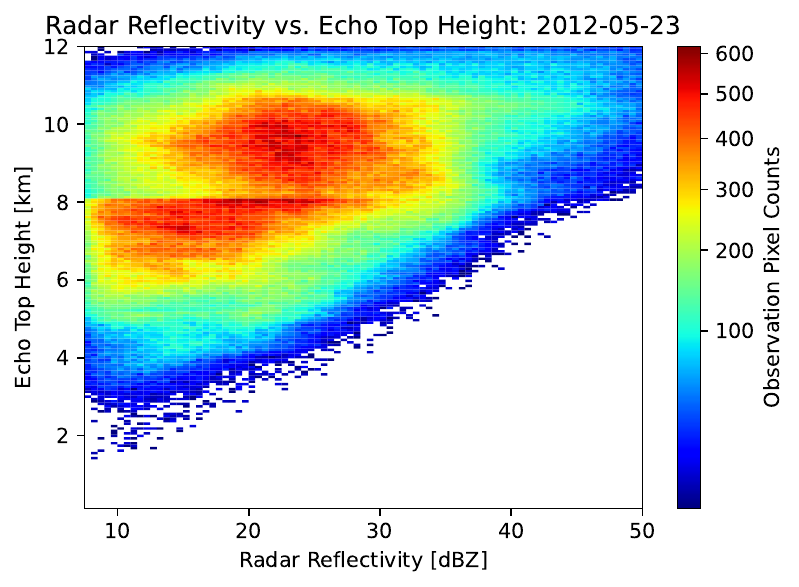}
    \includegraphics[width=0.49\linewidth]{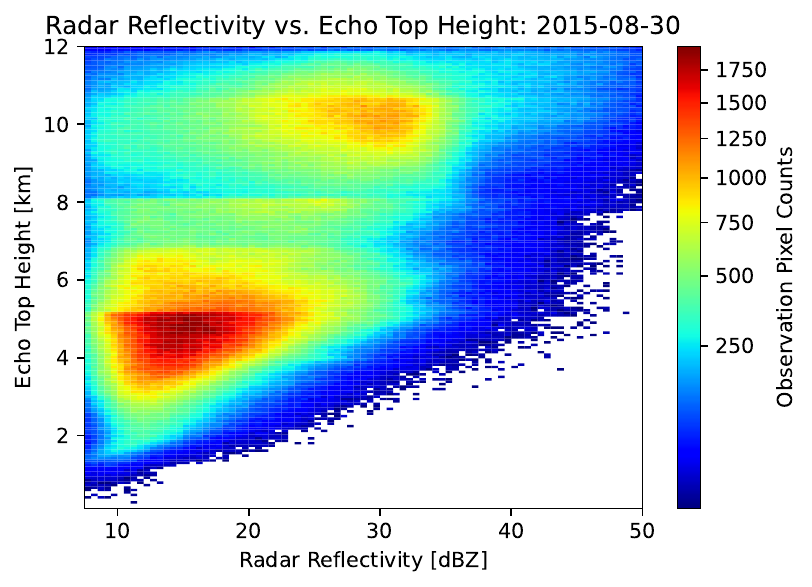}
    \caption{The distribution of radar reflectivity with their corresponding echo top height observations for three different days with strong precipitation. The plots suggest a roughly proportional relationship between the two variables. The echo top data also clearly contains artifacts corresponding to the maximum observable echo top height of various radar sweeps, visible as density cutoffs at various heights.}
    \label{fig:hist2d}
\end{figure}

To illustrate the relationship of echo top heights and reflectivity, we plot the co-occurrences of radar reflectivity and echo top height pairs of observations from three days of interest in Figure~\ref{fig:hist2d} (we use data from De Bilt and Den Helder stations in Netherlands, see Section \ref{subsec:data} for details about the data used). A strong relationship is visible in all of the plots -- the bottom right part is empty, meaning there are almost no observations with high rainfall rate and low echo top height. This implies that high values of echo top height could help us identify the extreme precipitation events -- as one variable increases, the other also increases proportionally. However, the correlations do not seem to be so strong that adding the echo top height variable would be meaningless.

\subsection{Related Uses of Echo Top Heights}

Echo top height (ETH) has served as a valuable resource in many applications. Higher ETH values generally indicate stronger updrafts and more intense convection, often associated with severe weather phenomena like heavy precipitation, hail, and lightning. In aviation, ETH is a critical parameter for assessing potential hazards to aircraft~\cite{hwang2015improved}. Furthermore, ETH has been investigated for its utility in predicting lightning initiation, with studies suggesting that ETH reaching at least 7 km are a necessary condition for cloud-to-ground lightning strikes~\cite{yang2010investigating}, showing a connection to lightning initiation during storms.

Notably, ETH has also been explored as an indicator of rainfall rate incorporated into quantitative precipitation estimation -- the task of mapping radar observations to the precipitation on ground. In~\cite{zou2023radar}, the use of ETH in conjunction with radar reflectivity (Z) for quantitative precipitation estimation using a GRU neural network was explored. This research demonstrated that the inclusion of ETH as an input feature significantly improved the accuracy of rainfall estimation compared to using only radar reflectivity. Although this study concentrated on precipitation estimation rather than nowcasting, it provides compelling evidence for the value of ETH in a machine learning context related to precipitation analysis.

Likewise, it was shown that echo top height is one of the factors that can influence the relationship between radar reflectivity (Z) and rainfall rate (R)~\cite{wu2018dynamical}. This suggests that incorporating ETH directly into a nowcasting model might help to implicitly account for the variability in the Z-R relationship, potentially leading to more accurate rainfall predictions.

\subsubsection*{Section Summary}

Based on the reviewed material, there is a noticeable gap in the current scientific literature regarding the direct utilization of echo top heights (ETH) as an input channel within deep learning models for precipitation nowcasting. The potential benefits of incorporating ETH warrant further investigation.

It is plausible that including ETH could enhance the model's ability to represent the vertical structure and intensity of convective systems more effectively than using reflectivity data alone. Changes in ETH over time might also provide valuable cues for predicting the initiation, growth, and decay of precipitation, particularly in convective events. Furthermore, its correlation with intense convection suggests it could improve the nowcasting of heavy precipitation. However, it it is possible that the information provided by ETH is not beneficial and there is a significant degree of redundancy when used alongside the radar reflectivity data.

In conclusion, the analysis of the provided scientific literature indicates that leveraging Echo Top Heights in deep learning models for precipitation nowcasting represents a promising, yet relatively underexplored, research avenue.

\section{Methodology}

To evaluate the potential benefits of incorporating echo top height (ETH) in deep learning-based precipitation nowcasting, a suitable neural network architecture must be selected. While recent work in the field is increasingly focused on generative approaches, these models are computationally expensive to train and evaluate. Moreover, achieving state-of-the-art performance is not the primary aim of this study; rather, our focus is on proof-of-concept experimentation.

Given these considerations, we opt for a deterministic model that produces a single nowcast rather than a distribution of possible outcomes. Although deterministic models are affected by the so-called double penalty problem -- manifesting as spatial blurring and intensity degradation over time -- we consider this a potential advantage in our context. If the inclusion of ETH effectively reduces uncertainty in the predicted evolution of precipitation, any improvement should be visible as a reduction in this blurring effect.

We specifically selected the U-Net architecture as the neural network architecture experiment for it's simplicity and wide use in nowcasting. See Section~\ref{subsec:arch} for more details on the U-Net architecture used and the training setup. We use mean squared error as a loss function to minimize.

To ensure our experimental results are not due to random chance, we train multiple models using different random seeds and varying train-validation splits. This helps assess the consistency and robustness of the models across different initializations and data subsets. In addition to visual, qualitative evaluation on selected nowcasting events, we use a combination of general pixel-wise metrics, threshold-based metrics over binarized maps and also include the Fractions Skill Score (FSS), a very popular metric used in forecasting which evaluates spatial accuracy at different scales and accounts for slight spatial misalignments in precipitation fields~\cite{roberts2008scale}.

\subsection{Dataset} \label{subsec:data}

The data used for training and validating the model consists of radar observations for the Netherlands provided by the KNMI~\cite{knmi_data_platform}. The reflectivity data can be downloaded from the KNMI data platform at this link~\cite{knmi_radar_refl_2024} and the corresponding echo top heights here~\cite{knmi_echotopheight_2024}. The datasets were generated by combining the data from two C-band radars in De Bilt (now in Herwijnen) and Den Helder at various elevation angles. For the 1500~m reflectivity product, a few low-elevation angle scans are interpolated and combined using a distance-weighted average. For echo top height products, more elevation angles are used to find the highest altitude where reflectivity exceeds 7~dBZ, and the maximum value from both radars is taken. In both cases, the file timestamp reflects the start of the lowest scan and the data are updated at regular 5 minute intervals.

The whole dataset spans a 15-year period  between 2008 and 2022, comprising over 1.5 million observations at a 5-minute temporal resolution, excluding occasional missing data. The majority of these observations represent clear-sky conditions, which, if included, would introduce a significant bias toward low-precipitation events. To address this imbalance, we computed a precipitation event weight for each observation by summing the squared radar reflectivity values across all pixels. Then, we selected the top 1000 observations from each year and used them as the starting points of dataset sequences. Each training sequence consists of 22 consecutive observations, which results in 15000 overlapping sequences consisting of 24118 total selected observations. In total, only around 1.5~\% of the total available data were included in the dataset for training and evaluating the model.

However, during preprocessing, we identified a significant distributional shift in the echo top height data beginning in the second half of 2016, coinciding with the installation of new polarimetric radars by KNMI. Therefore, to ensure data consistency, we restricted model training to data collected from October 2016 onward.

\begin{figure}[!h]
  \centering
  \includegraphics[width=0.66\linewidth]{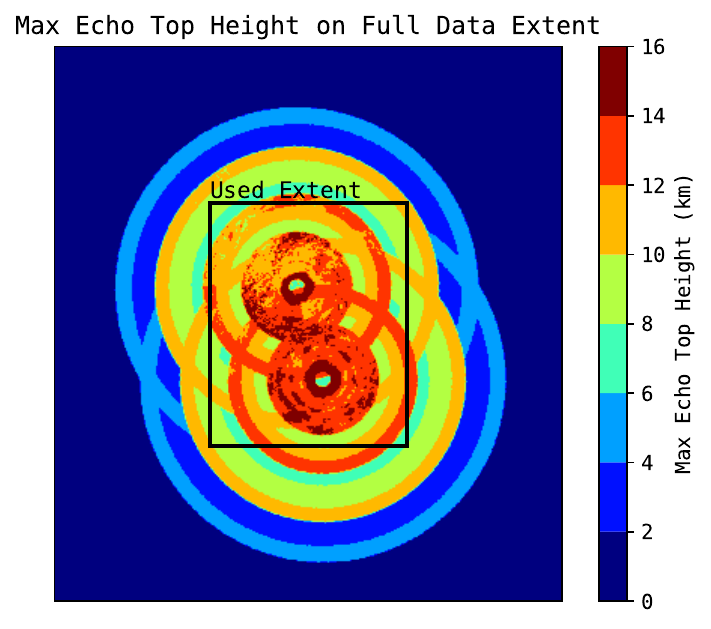}
  \caption{Maximum observed echo top height per pixel across all training and test set samples (October 2016 to December 2022). The highlighted area indicates the spatial extent selected for model training, ensuring vertical coverage of at least 7 km.}
  \label{fig:ETH_Max}
\end{figure}

In terms of spatial resolution, each pixel in the dataset corresponds to a $1\times1$ km area. The full radar image spans $765\times700$ km. However, a significant portion of the image lies beyond the effective range of the radars. This limitation becomes particularly evident in the echo top height data. Since the radar performs sweeps at multiple elevation angles, the maximum detectable altitude varies across the image. Near the edges of the coverage area, which is only reached by the lowest-angle sweeps, the radar cannot detect echo tops above approximately 2 km. In contrast, the echo top heights at the center of the image can reach up to 16 km. To ensure consistent vertical coverage across the input data, we restricted the training dataset to a central subset of the image ($336\times272$ km), where echo tops of at least 7 km are detectable across the entire region. The final spatial extent used for training is shown in Figure~\ref{fig:ETH_Max}. While this still results in considerable artifacts present in the echo top data when objects travel between sweeps, a smaller extent would be undesirable as it would drastically reduce the spatial coverage and amount of training data.

For the model training and evaluation, we convert the radar reflectivity to rainfall rate using the using the Marshall-Palmer formula~\cite{THEDISTRIBUTIONOFRAINDROPSWITHSIZE}. To mitigate speckle-like noise in the radar reflectivity data, we apply a morphological clutter removal procedure to each frame of the radar composite, following the approach used in~\cite{vanos2024precipitation}. First, the reflectivity field is thresholded to create a binary mask, where pixels corresponding to rain rates above 0.1 mm/h are set to 1, and all others to 0. This binary mask is then processed using a morphological opening operation, consisting of three iterations of erosion followed by three iterations of dilation. This procedure effectively removes small, isolated reflectivity regions that are likely the result of measurement noise or non-precipitating clutter.

As a test set for evaluation, we use the sequences from the last year of the dataset -- 2022. Besides using these for quantitative evaluation, we also picked three precipitation events of interest for qualitative evaluation purposes in Section~\ref{sec:qualitative}.

\subsection{Architecture Exploration and Preliminary Experiments} \label{subsec:arch}

We conducted a series of preliminary experiments to explore how echo top height (ETH) data could be integrated into deep learning-based precipitation nowcasting. As a foundation, we implemented a variant of the RainNet U-Net architecture proposed by Ayzel et al.~\cite{ayzel2020rainnet}, which was extended in several directions to assess the effect of ETH and different modeling strategies.

To incorporate ETH, we first experimented with a straightforward extension of the original 2D convolutional model by interleaving ETH and rainfall observations as additional input channels. However, this conflated temporal, and variable dimensions. To address this, we restructured the input to treat ETH and rainfall as separate channels across time, leading to a natural use of 3D convolutions in the U-Net (where time is the third dimension) that could now model spatial and temporal dependencies jointly.

Early results suggested that both the inclusion of ETH and the shift to 3D convolutions improved short-term prediction accuracy. However, we observed increasing systematic bias at longer lead times in models since they used recursive input strategy where past predictions feed into future forecasts.

To mitigate this, we tried constraining ETH outputs during training by incorporating a secondary loss term. While this reduced bias accumulation, it also degraded overall forecast performance. As an alternative, we modified the architecture to predict all future frames in a single forward pass, removing the recursive loop. We chose to predict 18 future frames at once, resulting in nowcasts from 5 to 90 minutes into the future. This approach eliminated the long-term bias issue and maintained comparable accuracy to the iterative approach.

In short, initial experiments showed that incorporating ETH as an additional input and adopting 3D convolutions (treating time as a separate dimension) improved predictive accuracy. However, these models exhibited increasing systematic bias at longer lead times due to recursive input dependencies. Attempts to mitigate this bias by constraining ETH predictions during training reduced the bias but degraded overall accuracy. An alternative approach—removing the recursive strategy and instead directly predicting all future frames—eliminated the bias and maintained competitive accuracy.

\subsection*{Section Summary}

The study opts for a deterministic, U-Net-based architecture for simplicity and interpretability, using mean squared error as the loss function. Despite its susceptibility to blurring over time, the model's performance degradation is considered useful for identifying ETH's effect in reducing prediction uncertainty. The dataset comprises radar reflectivity and ETH data from the Dutch KNMI, spanning 2008–2022. However, due to changes in radar hardware in 2016, to ensure data consistency, only post-October 2016 data is used. The final model used in this study is a 3D U-Net variant that processes spatiotemporal data using 3D convolutions and predicts all 18 future precipitation frames in a single forward pass, avoiding recursive prediction and reducing long-term bias accumulation.

\section{Evaluation}

To ensure the robustness and reliability of our experimental results — and to rule out the possibility that they are due to random chance or favorable initial conditions - we trained eight instances of our models, both with and without echo top height inputs while using different random seeds as well as varying train-validation data splits. Each model used roughly one eighth of the training sequences as validation set for early stopping and learning rate scheduling. All the models were evaluated on the same test set, that is, sequences from the last year of the dataset in 2022.

Results are assessed both quantitatively and qualitatively. On the qualitative side, we visually inspected the forecasts on selected events to assess how well the different models capture spatial and temporal patterns in precipitation evolution, particularly in cases with heavy rain and complex rainfall dynamics. On the quantitative side, we use a combination of general-purpose pixel-wise error metrics and domain-specific verification scores. The pixel-wise metrics include Mean Absolute Error (MAE), which measures the average magnitude of errors regardless of direction; Mean Squared Error (MSE), which penalizes larger errors more severely and is sensitive to outliers; and Mean Error (ME), also known as bias, which indicates whether the model systematically over- or underestimates precipitation. Ta gain additional insight about performance at various precipitation levels, we further include precision, recall, and the equitable threat score (ETS), which jointly assess detection skill, false alarms, and overall forecast accuracy for selected binary precipitation thresholds. In addition to these, we also considered the Fractions Skill Score (FSS), a widely-used metric in the nowcasting community that evaluates spatial consistency and skill at different spatial scales. The FSS is particularly useful in precipitation forecasting as it accounts for slight spatial misalignments between predicted and observed fields—something traditional pixel-wise metrics tend to penalize heavily, even if the forecast is meteorologically reasonable.

\subsection{Quantitative Evaluation: MAE, MSE, ME}

\begin{figure}[!h]
    \centering
    \includegraphics[width=0.49\linewidth]{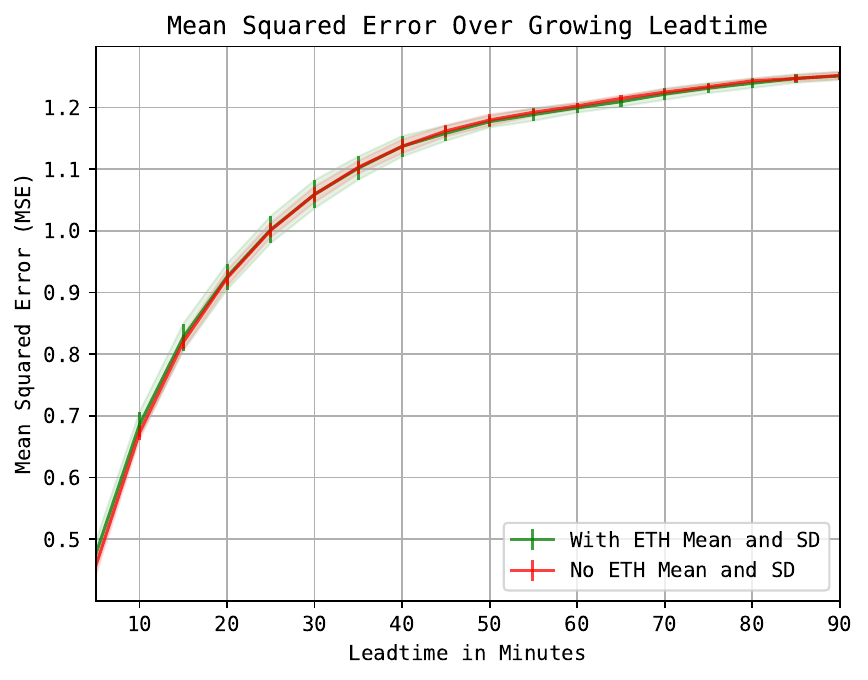}
    \includegraphics[width=0.49\linewidth]{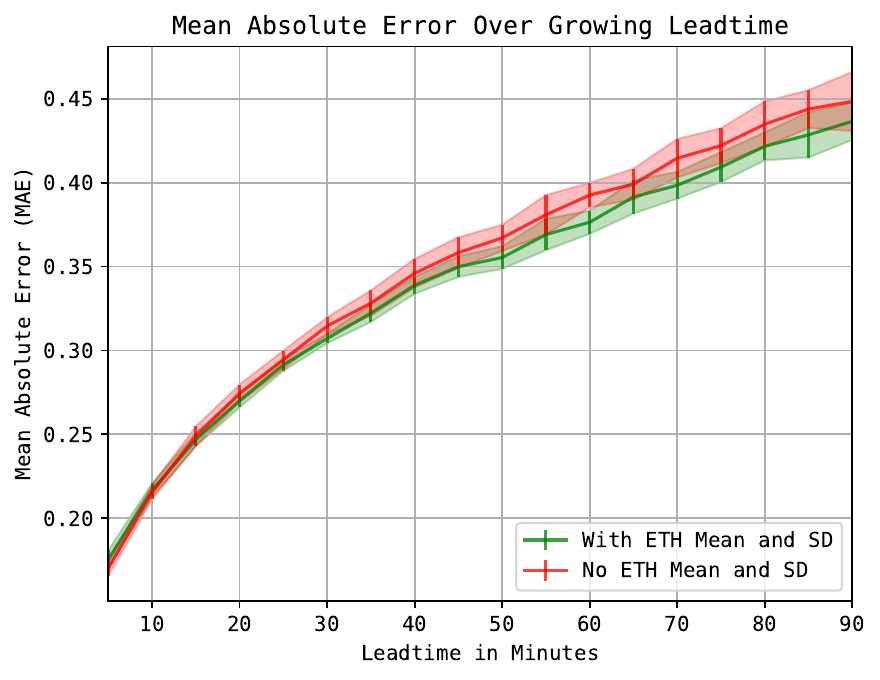}
    \includegraphics[width=0.49\linewidth]{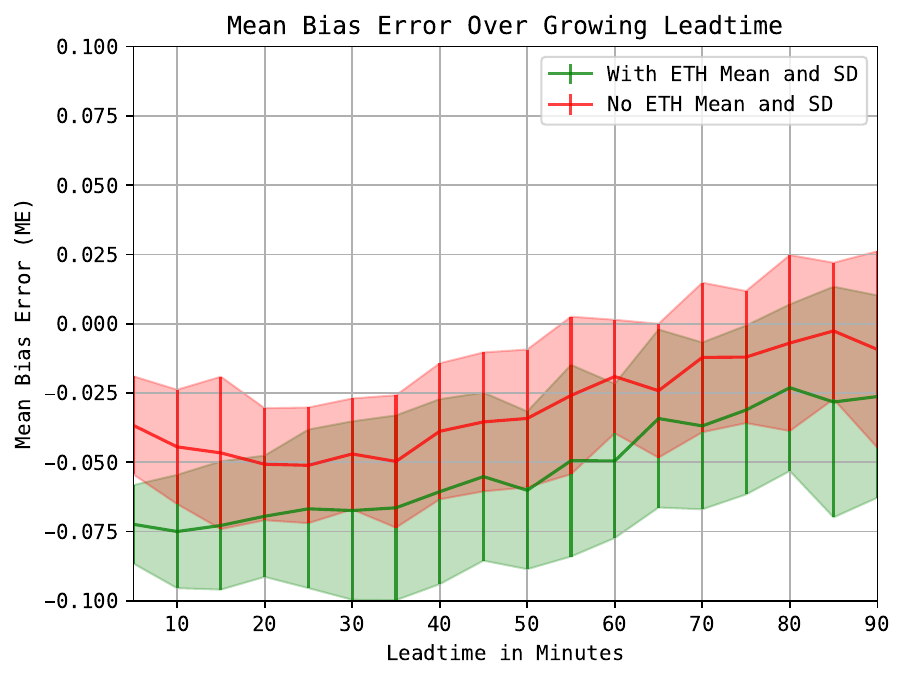}
    \caption{The means and standard deviations of models with and without echo top height input for three metrics -- mean squared error (MSE), mean absolute error (MAE), and mean error (ME) or bias. Each group consists of 8 separate models trained of with different test-validation splits and different random seeds for both the models with (green) and without (red) echo top heights.}
    \label{fig:8v8m}
\end{figure}

We begin by evaluating model performance using standard point-wise metrics: mean squared error (MSE), mean absolute error (MAE), and mean error (ME), also referred to as bias. To assess the variability across models, we computed the mean and standard deviation of each metric at all forecast lead times for two groups of models: eight trained with echo top height (ETH) input and eight without. This setup allows for a comparison of predictive accuracy and consistency between the two approaches. The aggregated results are presented in Figure~\ref{fig:8v8m}.

The MSE results indicate nearly identical average errors between the two groups, though the ETH-based models display greater variability. The MAE curves show that models with ETH consistently perform better at longer lead times, suggesting improved stability in the nowcasts. However, the ME plots reveal a more pronounced negative bias in the ETH-enhanced models, suggesting a systematic tendency to underestimate precipitation compared to their non-ETH counterparts.

The greater variance in MSE for the models with ETH suggests increased sensitivity. The outputs of the models that rely on ETH features vary more strongly across different training subsets, suggesting that performance is likely very case specific. Despite this, the improved MAE at longer lead times suggests that ETH provides useful contextual information, even if the absolute precipitation levels are occasionally underestimated.

\begin{figure}[!h]
    \centering
    \includegraphics[width=0.49\linewidth]{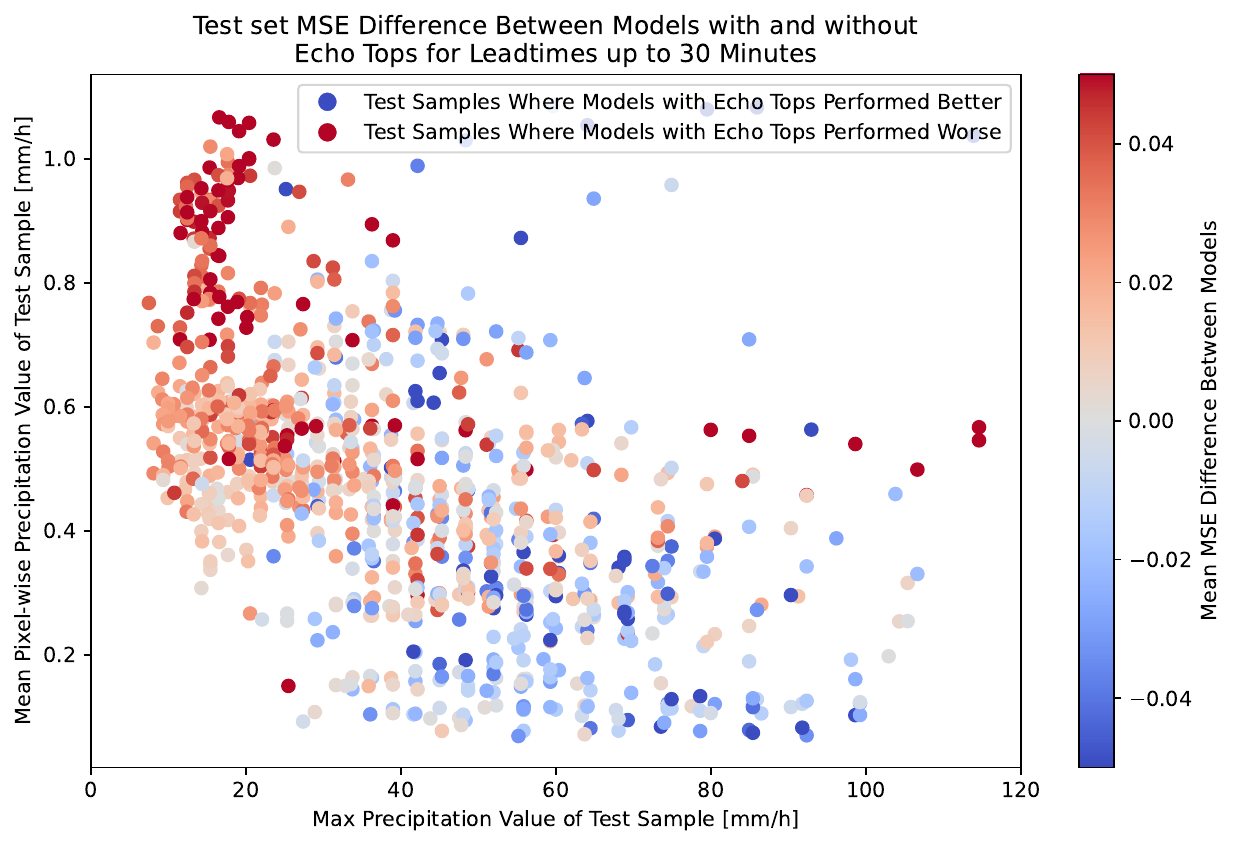}
    \includegraphics[width=0.49\linewidth]{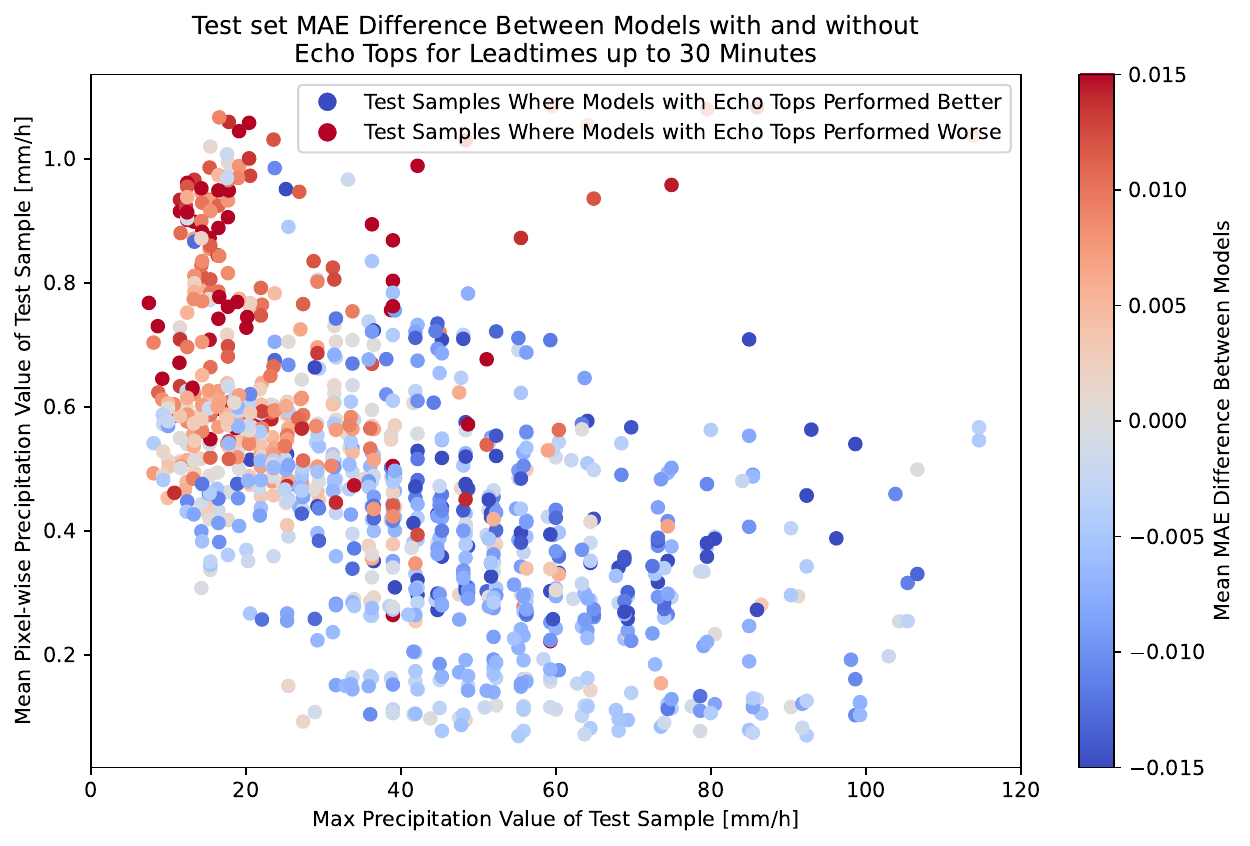}
    \includegraphics[width=0.49\linewidth]{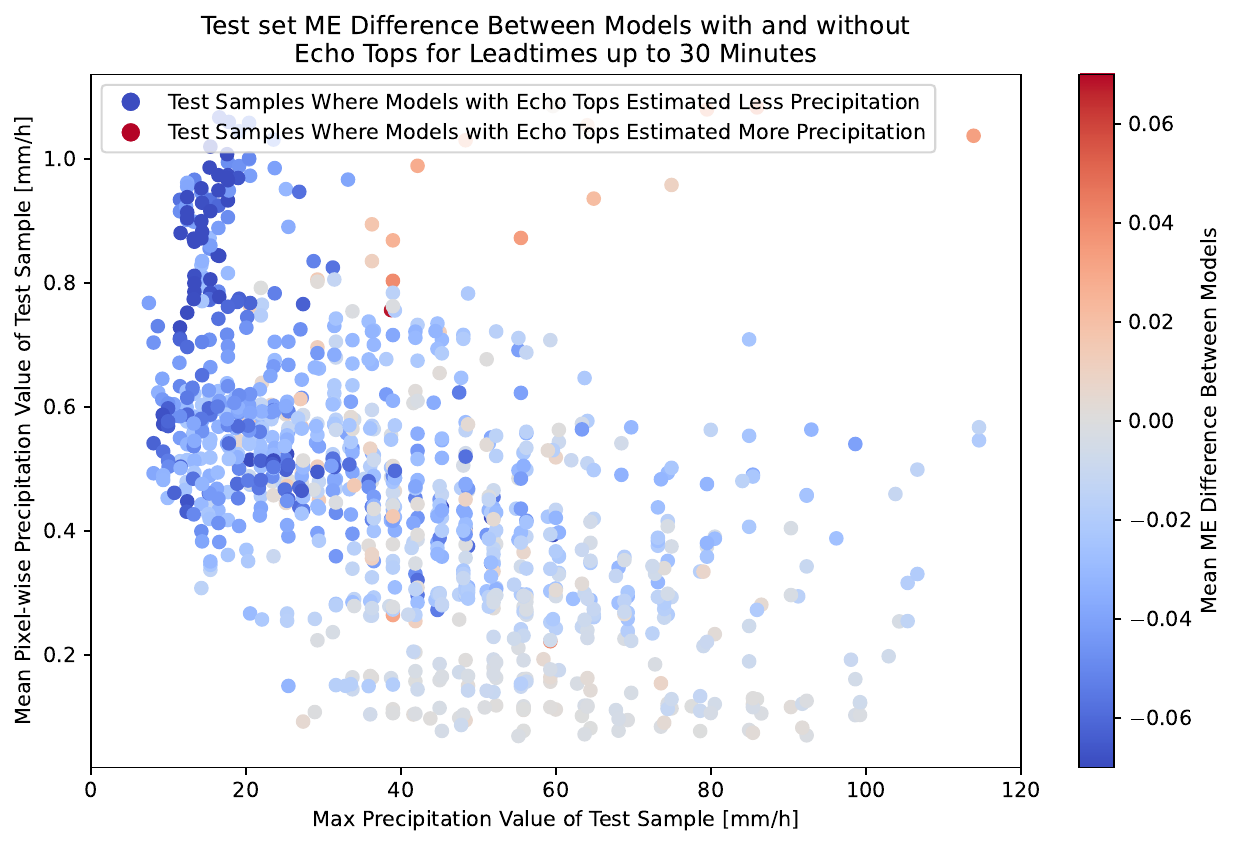}
    \caption{Mean difference of models with and without echo top height input for three metrics -- mean squared error (MSE), mean absolute error (MAE), and mean error (ME) or bias -- for each test sample based on their maximum and spatially averaged precipitation rate.}
    \label{fig:8v8diff}
\end{figure}

To further investigate the conditions under which the models perform better or worse, we analyzed the test set samples as a function of observed precipitation intensity. Specifically, each sample was ranked according to its maximum and spatially averaged precipitation rates at 30 minutes lead time, providing a two-dimensional representation of event severity. We then computed the mean difference in evaluation metrics between the two model groups with and without echo top height (ETH), and used this difference to color each sample accordingly. This visualization, shown in Figure~\ref{fig:8v8diff}, illustrates how the relative performance of ETH-enhanced models varies across different precipitation events.

The MSE and MAE plots reveal a distinct pattern: models with ETH input tend to perform worse on events with broad but moderate rainfall (i.e., higher average precipitation and lower peak intensities), yet perform better on events characterized by localized, high-intensity precipitation (i.e., higher maximum values and lower spatial averages). This suggests that ETH information helps the model focus on intense convective features but also introduces noise and uncertainty in more uniform rainfall conditions. The ME plot, on the other hand, shows no consistent relationship with either precipitation metric. The observed reduction in absolute bias for lower-rainfall events is expected, as such cases inherently limit the range of possible prediction errors.

\subsection{Quantitative Evaluation: Precision, Recall, ETS}

\begin{figure}[!h]
    \centering
    \includegraphics[width=0.32\linewidth]{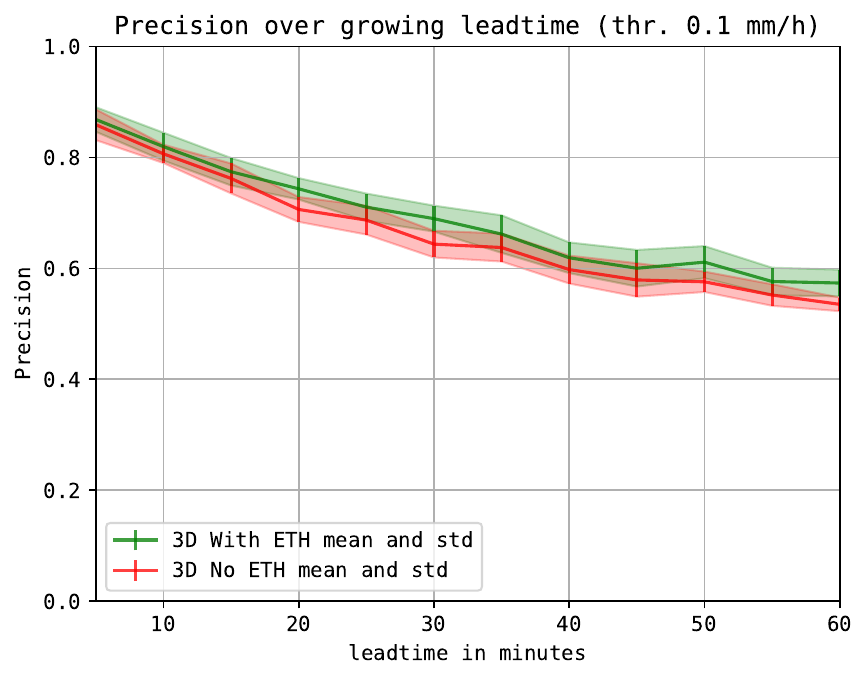}
    \includegraphics[width=0.32\linewidth]{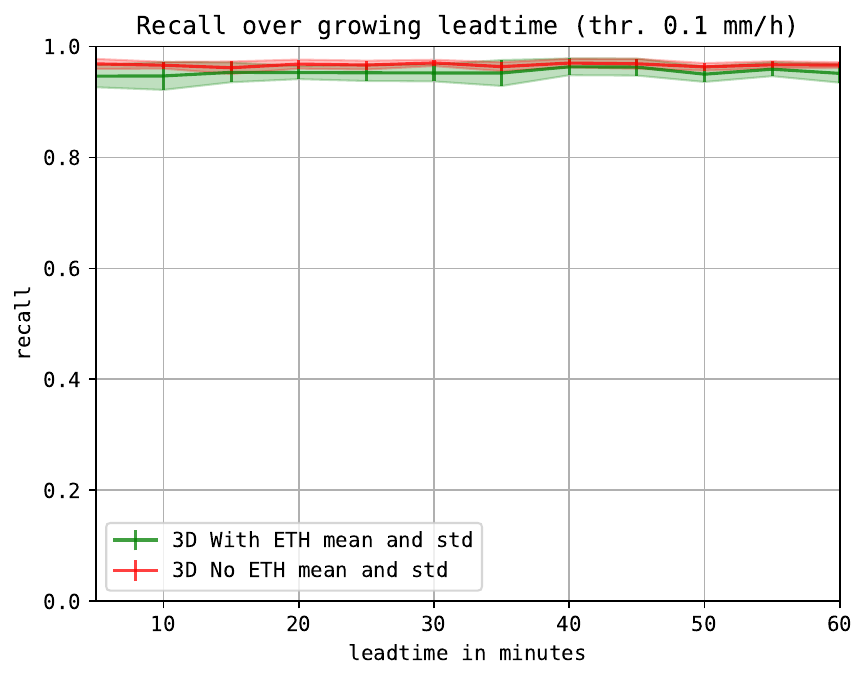}
    \includegraphics[width=0.32\linewidth]{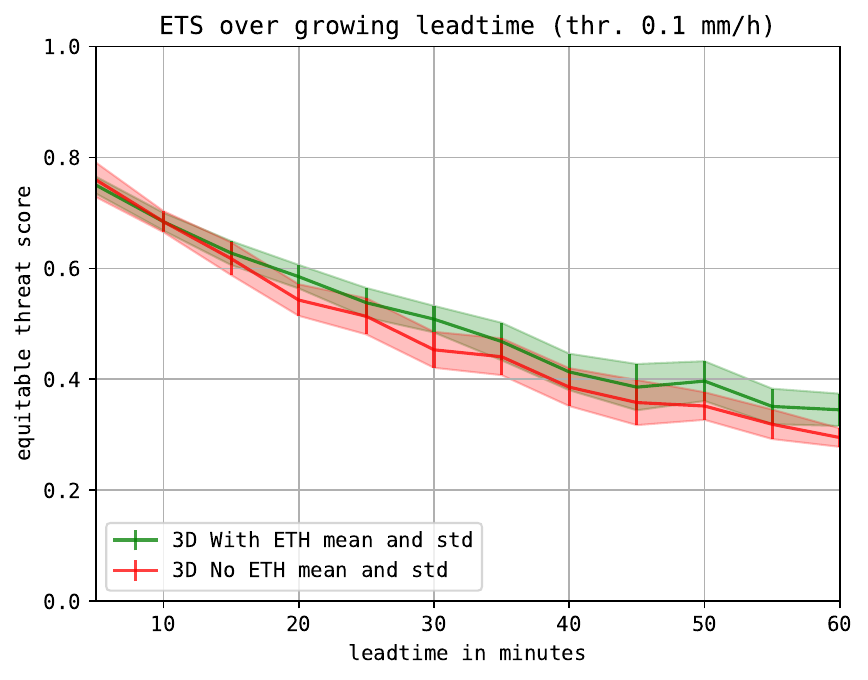}
    
    \includegraphics[width=0.32\linewidth]{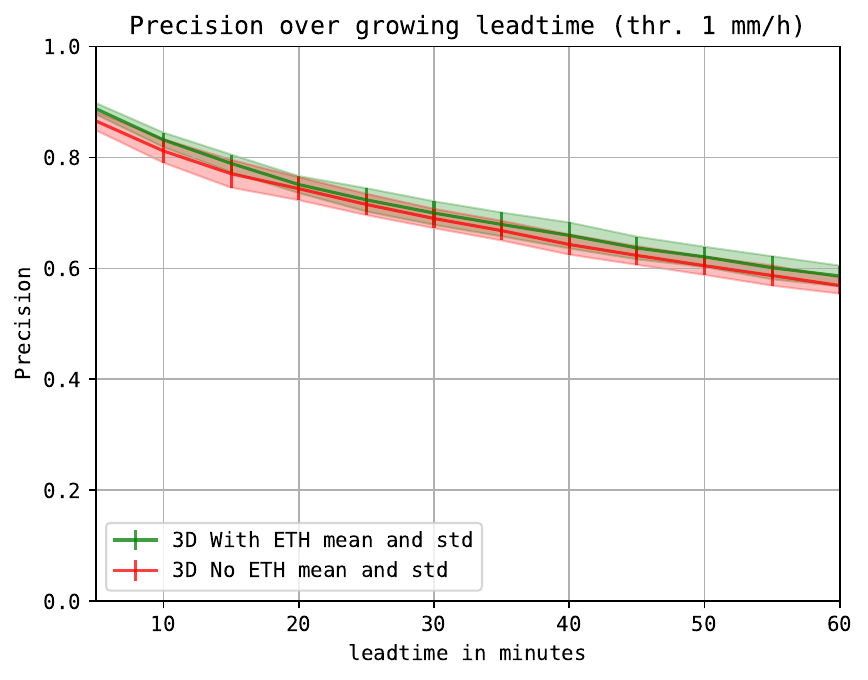}
    \includegraphics[width=0.32\linewidth]{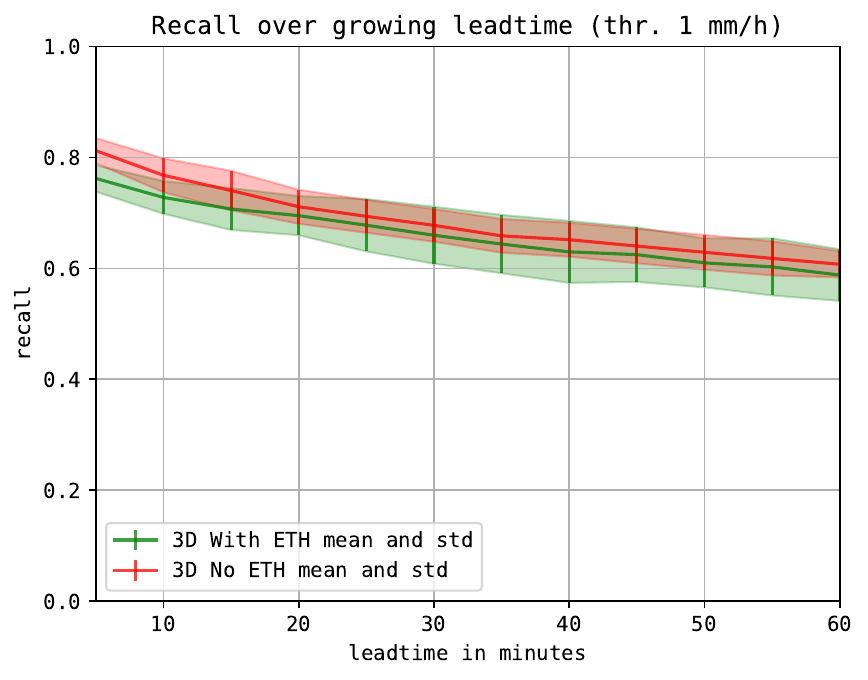}
    \includegraphics[width=0.32\linewidth]{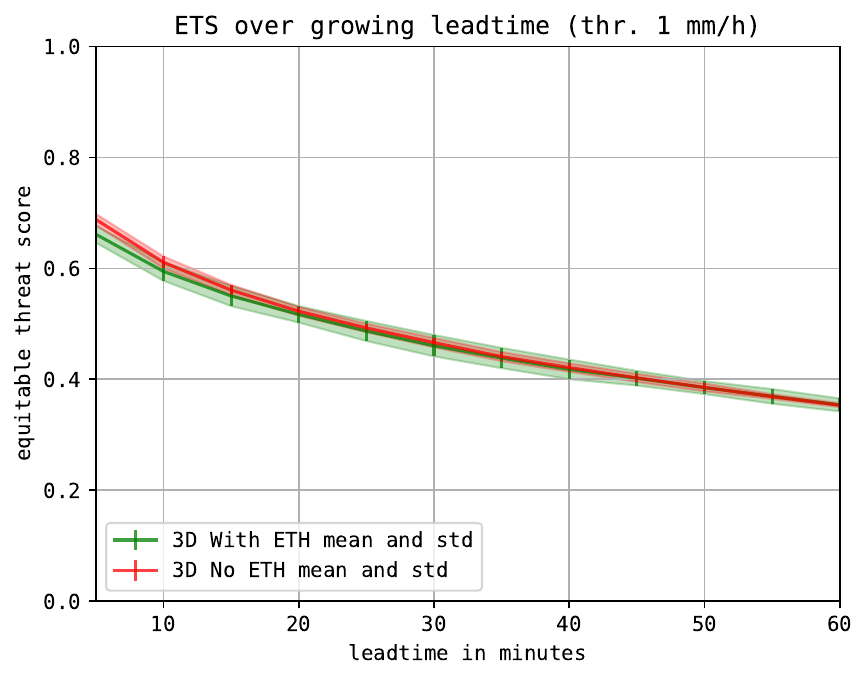}
    
    \includegraphics[width=0.32\linewidth]{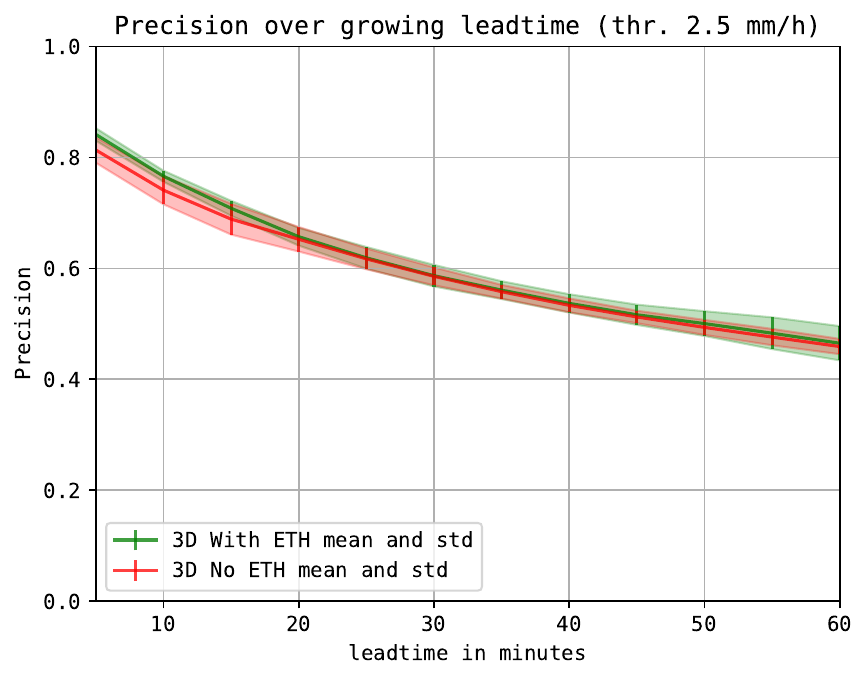}
    \includegraphics[width=0.32\linewidth]{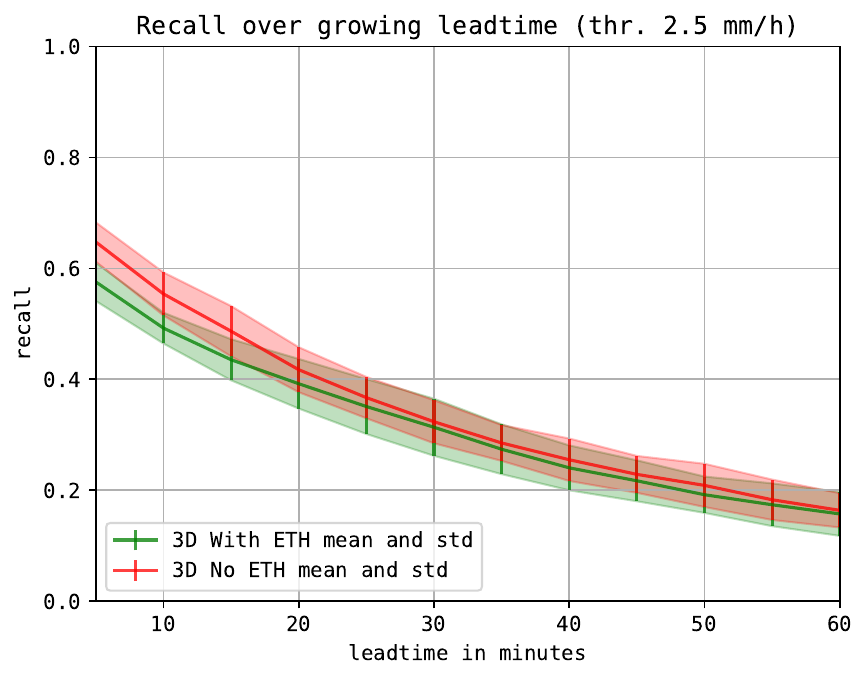}
    \includegraphics[width=0.32\linewidth]{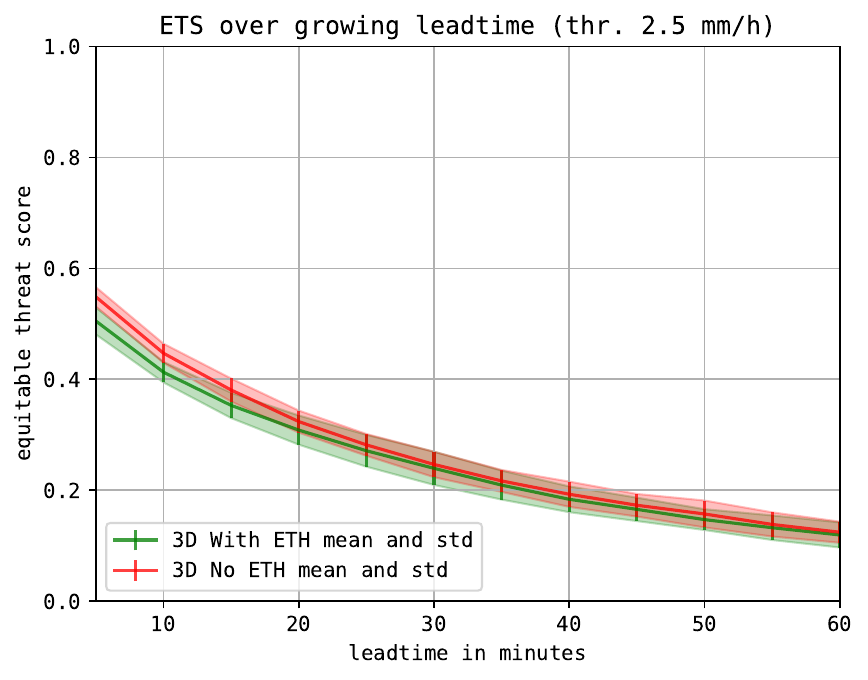}
    
    \includegraphics[width=0.32\linewidth]{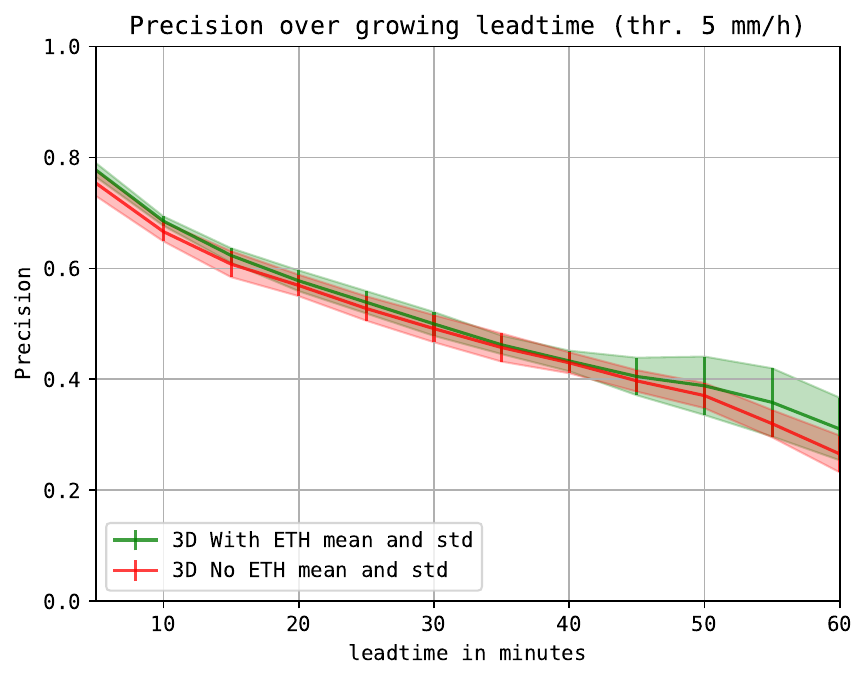}
    \includegraphics[width=0.32\linewidth]{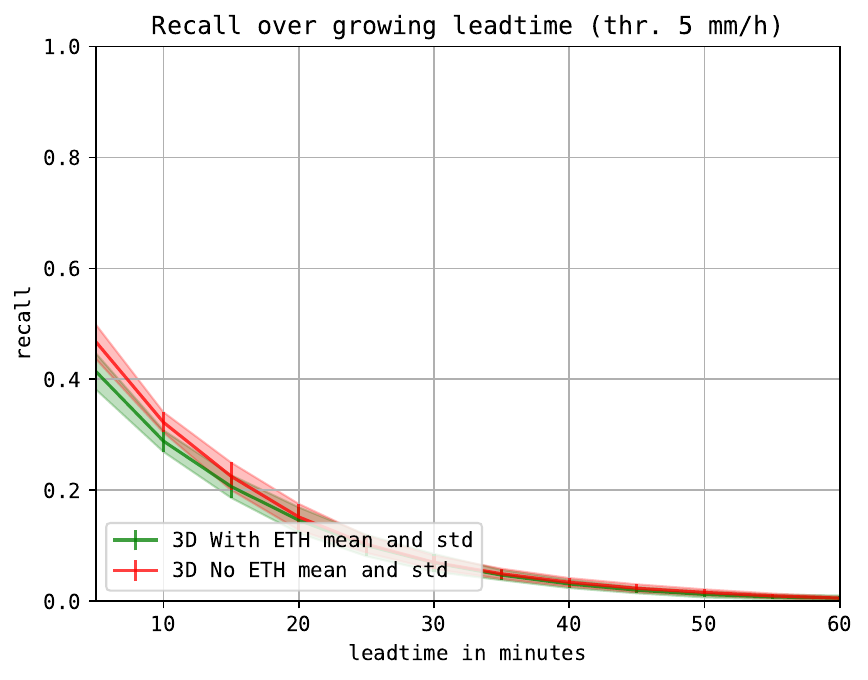}
    \includegraphics[width=0.32\linewidth]{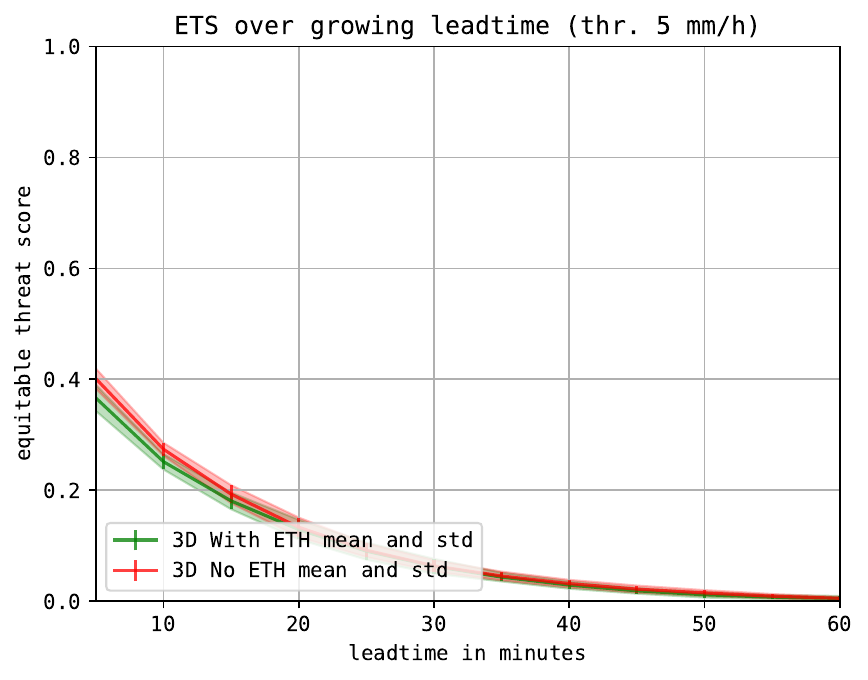}

    \caption{Means and standard deviations of models with and without echo top height for three metrics -- precision, recall, and equitable threat score (ETS). Evaluations are performed at four rainfall intensity thresholds: 0.1, 1, 2.5, and 5~mm/h. Each group consists of 8 separate models trained with different test-validation splits and different random seeds. Green colors are used for the models with ETH and red colors for the models without ETH.}
    
    \label{fig:thresholded}
\end{figure}

To complement the pixel-wise error metrics and gain a more event-focused perspective on model performance, we evaluate the predictions using threshold-based categorical metrics: precision, recall, and equitable threat score (ETS). These metrics are particularly useful in assessing the ability of the models to correctly identify precipitation occurrences above specific intensity levels, which is critical for operational forecasting and decision-making. The evaluation is conducted using four precipitation rate thresholds -- 0.1, 1, 2.5, and 5 mm/h -- chosen to span a range from light drizzle to moderate rainfall. At each precipitation level, predicted and observed precipitation fields are thresholded into binary maps of 0 and 1, and the metrics are computed accordingly. Precision reflects the fraction of predicted precipitation events that were actually observed, recall measures the proportion of observed events that were correctly predicted, and ETS provides a balanced assessment that accounts for random hits. This thresholded analysis allows for a more nuanced understanding of how the inclusion of echo top height input affects the models' capacity to predict precipitation with varying intensity. See Figure~\ref{fig:thresholded} for the plots.

Models that utilize ETH input consistently achieve higher precision, particularly at lower thresholds, indicating a tendency of the models toward predicting less rainfall with fewer false alarms in light precipitation scenarios. Conversely, models without ETH input exhibit higher recall across all thresholds, reflecting a greater ability to detect precipitation events, though at the cost of an increased rate of false positives. The ETS, which balances precision and recall, favors ETH-based models at the lowest threshold (0.1~mm/h). However, at higher thresholds, models without ETH input demonstrate superior performance. These findings suggest that ETH information enhances the detection of weak precipitation signals but is less beneficial -- or even detrimental -- when trying to forecast higher rainfall intensities.

\subsection{Quantitative Evaluation: Fractions Skill Score}

\begin{figure}[!h]
    \centering
    \includegraphics[width=0.48\linewidth]{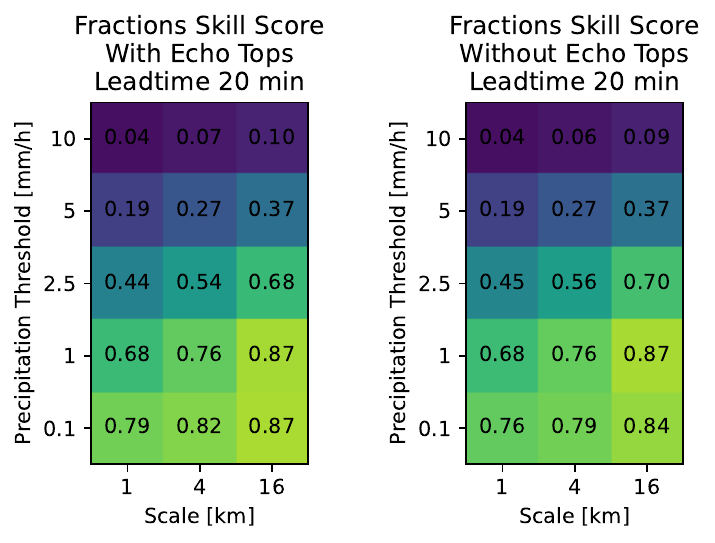}
    ~
    \includegraphics[width=0.48\linewidth]{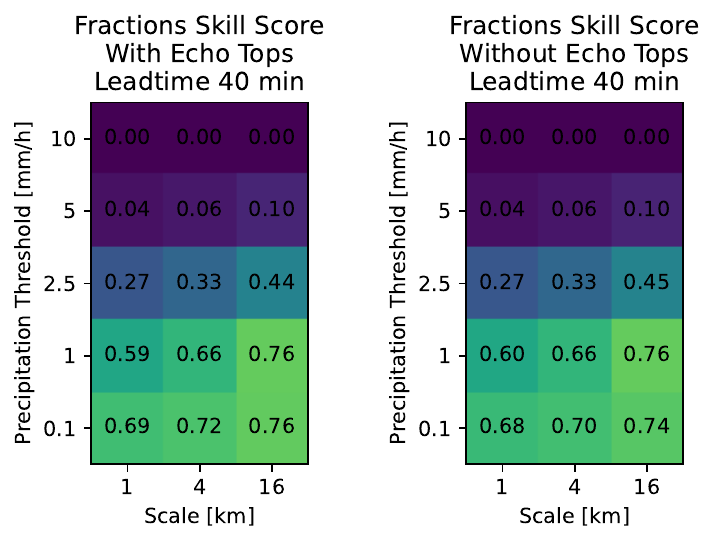}
    \includegraphics[width=0.48\linewidth]{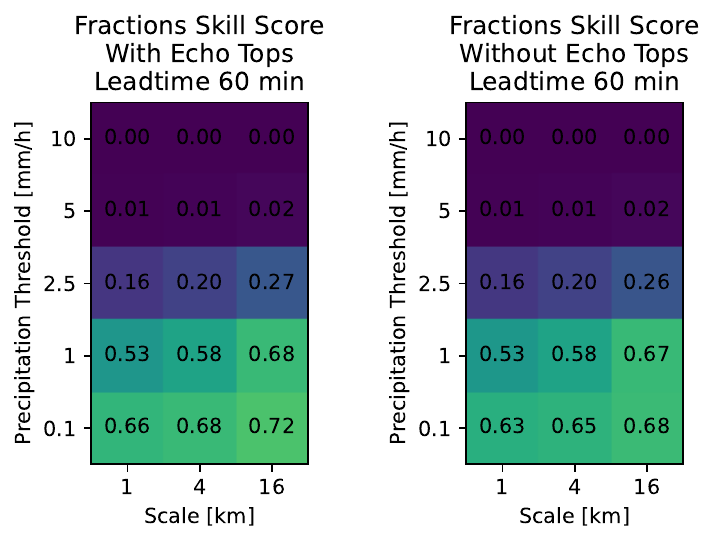}
    ~
    \includegraphics[width=0.48\linewidth]{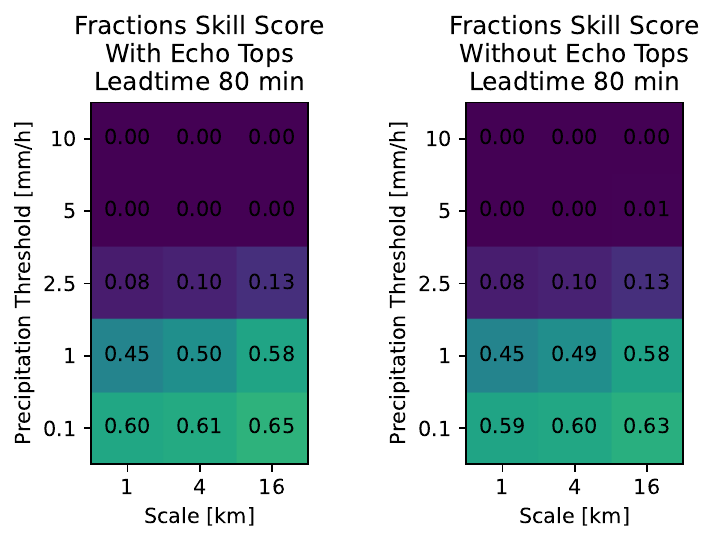}
    \caption{The matrices show the mean Fraction Skill Score (FSS) for models trained with and without echo top height (ETH) input. Evaluations are performed at five rainfall intensity thresholds (0.1, 1, 2.5, 5, and 10 mm/h) and three spatial tolerance scales (1, 4, and 16 km), across lead times of 20, 40, 60, and 80 minutes.}
    \label{fig:fss}
\end{figure}

To evaluate the spatial accuracy of predicted precipitation fields, we compute the Fractions Skill Score (FSS), a widely used metric in quantitative precipitation forecasting that accounts for both the intensity and spatial displacement of forecasted rainfall. Unlike pointwise metrics, FSS quantifies how well the predicted and observed precipitation fields align over neighborhood regions, making it suitable for assessing convective-scale forecasts where slight spatial shifts can lead to large pointwise errors. The FSS is evaluated across five precipitation rate thresholds (0.1, 1, 2.5, 5, and 10 mm/h) to capture performance across a range of rainfall intensities. Additionally, to examine the effect of spatial scale, the scores are computed using three spatial tolerance radii corresponding to 1, 4, and 16 km. This multiscale, thresholded approach provides a nuanced understanding of the models' ability to capture both occurrence of precipitation events across different intensities and lead times. See Figure~\ref{fig:fss} for the score matrices.

The results show that models incorporating echo top height (ETH) inputs achieve higher FSS values at lower precipitation thresholds, particularly for shorter lead times. This improvement is most evident at the 0.1 mm/h threshold, where ETH-based models consistently outperform those without ETH across all spatial scales. However, as the precipitation threshold increases, the differences in FSS between the two model configurations quickly diminish. At moderate to high thresholds (1~mm/h and above), the scores are similar, suggesting that the inclusion of ETH does not improve the spatial predictive skill for heavier rainfall events. From this, we conclude that the additional information provided by ETH mostly contributes toward improving the spatial coherence of light precipitation forecasts, while its benefit is limited for more intense rainfall.

\subsection{Qualitative Evaluation: Extreme Event Case Studies} \label{sec:qualitative}

Finally, we present qualitative examples of model outputs to complement the quantitative evaluation. As previously discussed, numerically assessing nowcasting performance is inherently challenging, as the optimal forecast often depends on specific application needs and may be subjectively interpreted. To gain further insight into model behavior, we selected three test samples for more in-depth visual analysis and comparison of the forecasts generated by models with and without echo top height input.

The selected events include:
\begin{itemize}
    \item \textbf{February 20, 2022:} Following closely after Storm Eunice, Storm Franklin brought strong winds and heavy rainfall, exacerbating damage from previous storms and causing localized flooding.
    \item \textbf{May 19, 2022:} On May 19, severe storms impacted the Netherlands, particularly affecting fruit farms in the Batavia region, leading to significant financial losses for fruit growers. Additionally, the southern province of Noord-Brabant and the city of Utrecht experienced flooding in homes and streets.
    \item \textbf{August 17, 2022:} On August 17, severe thunderstorms brought heavy rainfall to northeastern parts of the Netherlands, flooding streets and basements.
\end{itemize}

\begin{figure}[!h]
    \centering
    \includegraphics[width=\linewidth]{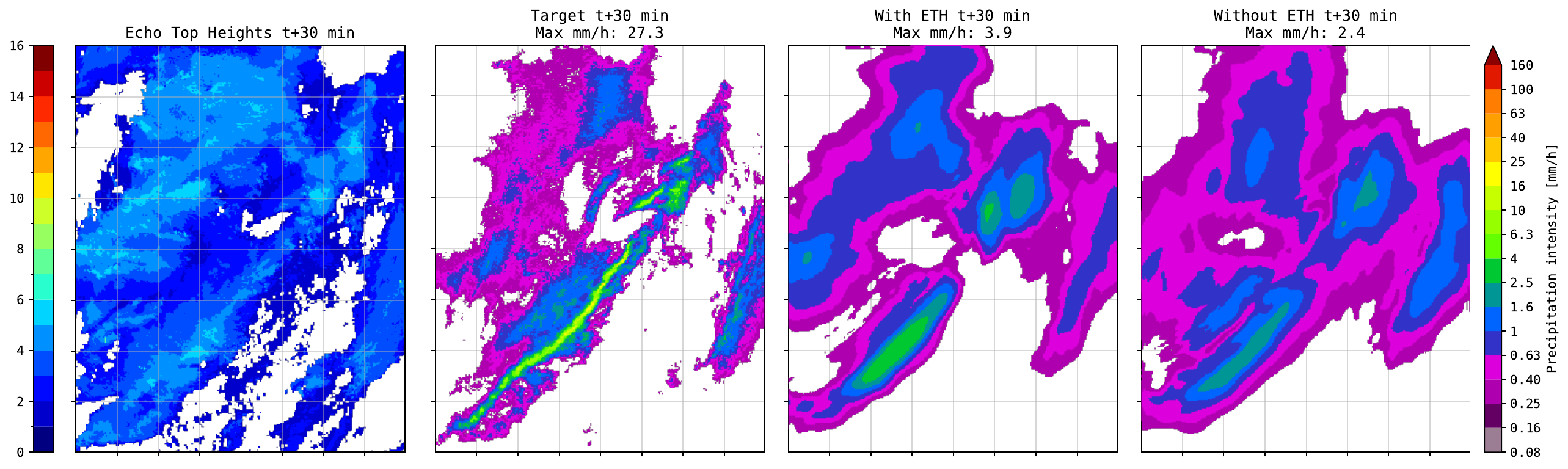}
    \includegraphics[width=\linewidth]{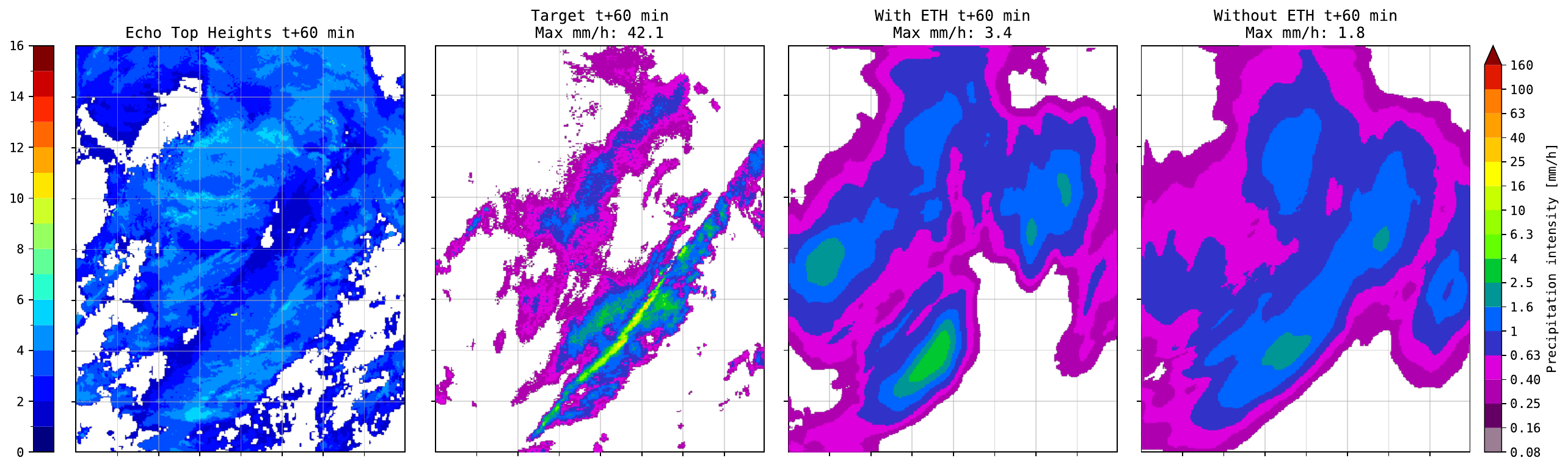}
    \includegraphics[width=\linewidth]{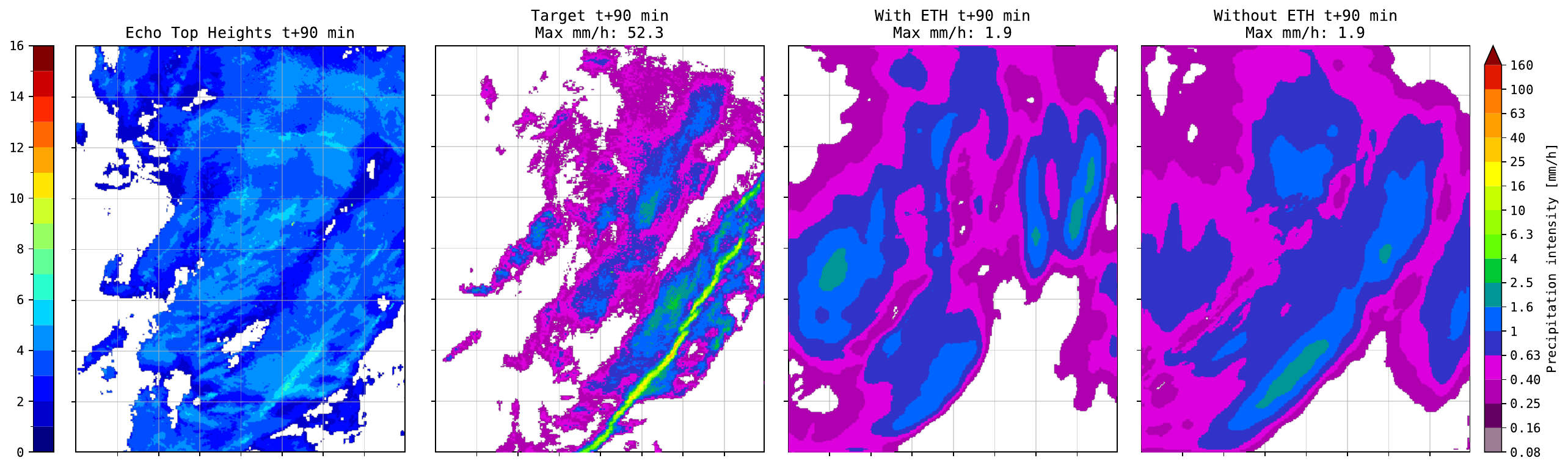}
    \caption{Series of nowcasts generated by the models with and without ETH for a precipitation event on February 20th, 2022, at 20:00. For reference, the corresponding ETH observation and ground truth rainfall rate observation are also shown. Each panel presents the predicted or observed rainfall rates at successive lead times (T+30, +60, +90 minutes). To help with interpreting intensity trends, the maximum rainfall rate within each map is indicated numerically above the corresponding panel, providing a quantitative measure of nowcast rainfall intensity degradation over time.}
    \label{fig:february}
\end{figure}

\begin{figure}[!h]
    \centering
    \includegraphics[width=\linewidth]{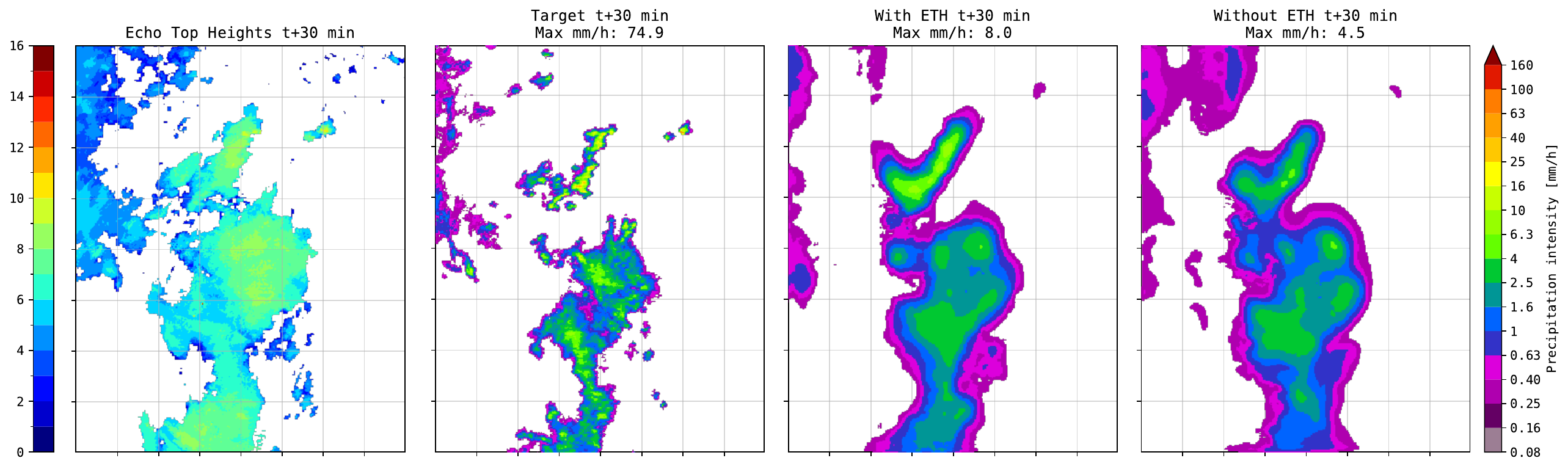}
    \includegraphics[width=\linewidth]{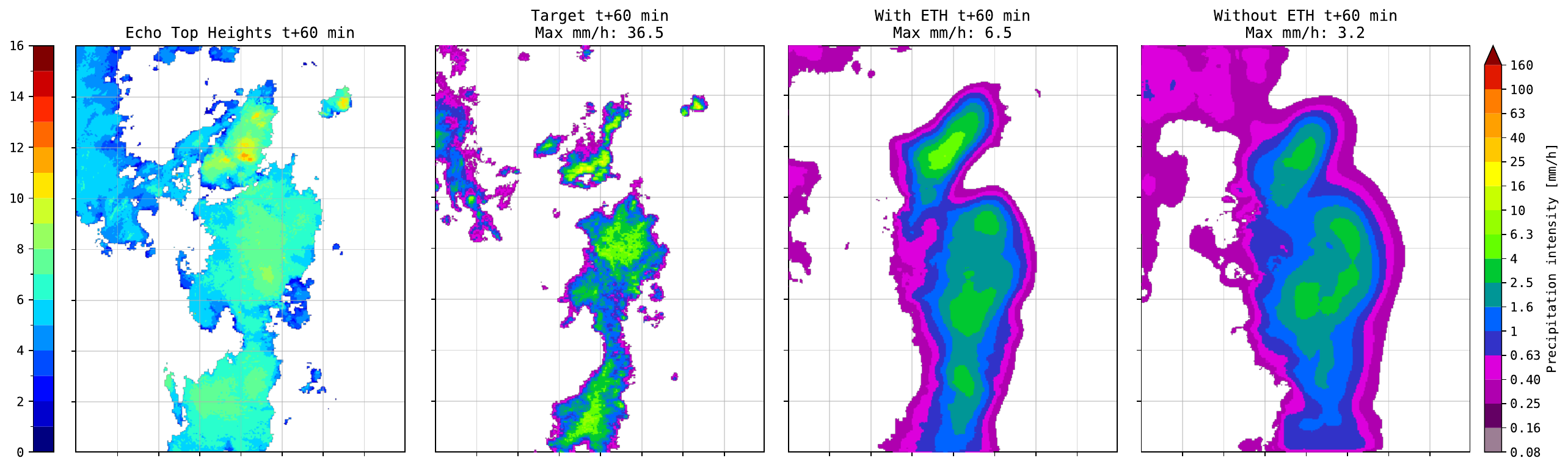}
    \includegraphics[width=\linewidth]{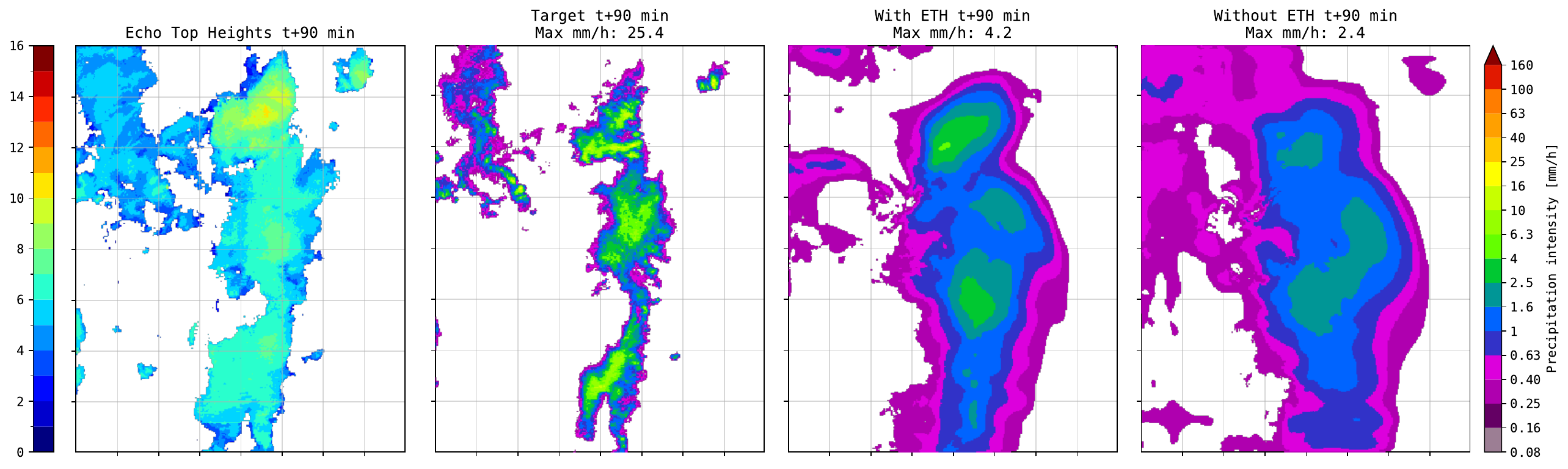}
    \caption{Series of nowcasts generated by the models with and without ETH for a precipitation event on May 19th, 2022, at 03:00. For reference, the corresponding ETH observation and ground truth rainfall rate observation are also shown. Each panel presents the predicted or observed rainfall rates at successive lead times (T+30, 60, 90 minutes). To help with interpreting intensity trends, the maximum rainfall rate within each map is indicated numerically above the corresponding panel, providing a quantitative measure of nowcast rainfall intensity degradation over time.}
    \label{fig:may}
\end{figure}

\begin{figure}[!h]
    \centering
    \includegraphics[width=\linewidth]{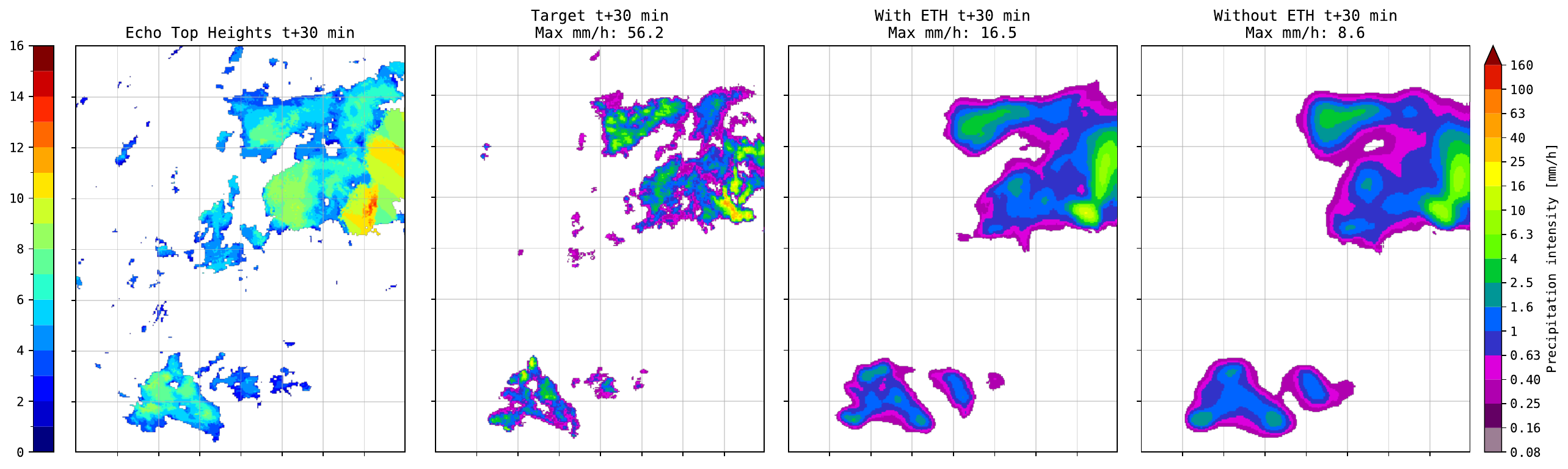}
    \includegraphics[width=\linewidth]{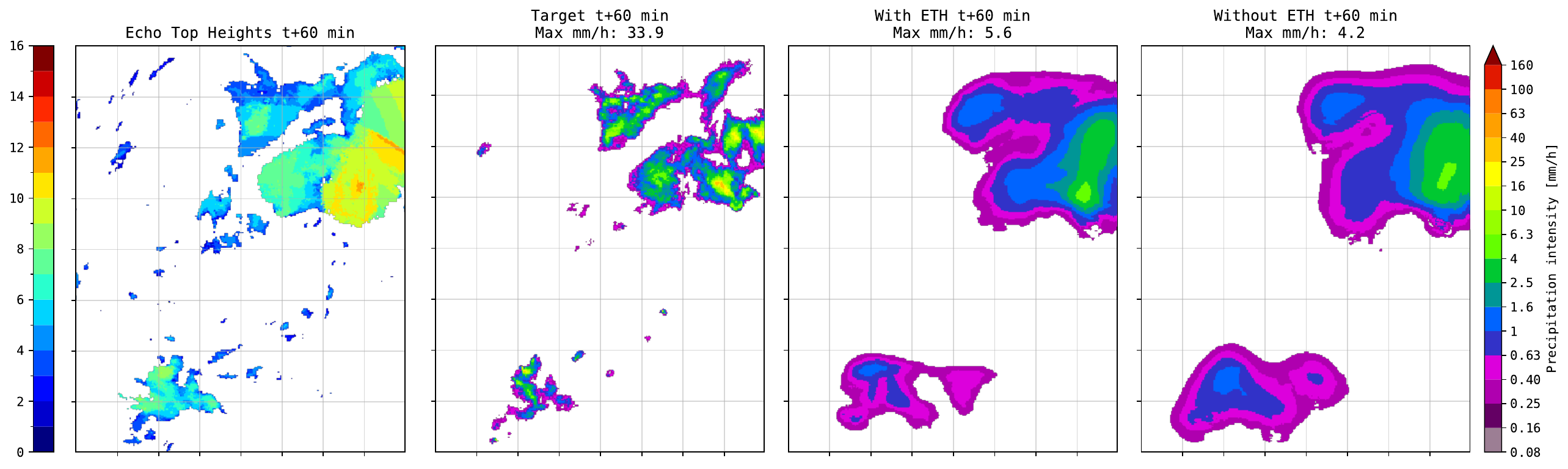}
    \includegraphics[width=\linewidth]{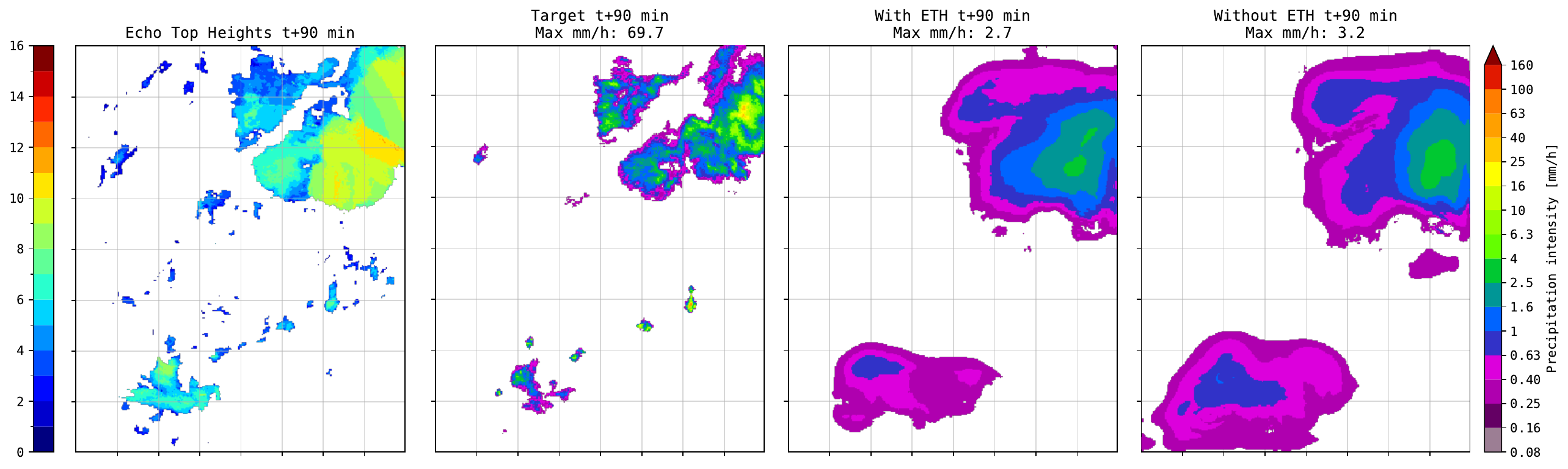}
    \caption{Series of nowcasts generated by the models with and without ETH for a precipitation event on August 17th, 2022, at 15:00. For reference, the corresponding ETH observation and ground truth rainfall rate observation are also shown. Each panel presents the predicted or observed rainfall rates at successive lead times (T+30, 60, 90 minutes). To help with interpreting intensity trends, the maximum rainfall rate within each map is indicated numerically above the corresponding panel, providing a quantitative measure of nowcast rainfall intensity degradation over time.}
    \label{fig:august}
\end{figure}

For the case studies above, we selected a single model from each group (i.e., with a without ETH) based on their performance over the entire test set. Specifically, we chose the model that achieved the highest fractions skill score (FSS) at a threshold of 2.5~mm/h, with a spatial tolerance scale of 16~km and a lead time of 30 minutes. As plotting each of the 16 trained model would be needlessly exhaustive, these were chosen to show the performance of "best" models of each group ploted side by side. The nowcasts generated by these selected models are visualized in Figures~\ref{fig:february}, \ref{fig:may}, and \ref{fig:august}.

Figure~\ref{fig:february} illustrates a case featuring a well-defined band of precipitation propagating eastward. Notably, although the rainfall intensity is high, the corresponding echo top heights (ETH) remain relatively low across the affected areas. When comparing the nowcasts, the model incorporating ETH input shows a modest improvement in preserving rainfall intensity over time. However, the model without ETH input appears to capture the overall spatial structure and propagation of the precipitation band better, maintaining the coherence of the main precipitation feature for longer.

Figure~\ref{fig:may} presents a case in which a region of intense precipitation, located near the center of the domain, propagates toward the northeast. In this instance, the associated echo top heights (ETH) are relatively high, exceeding 10 km in some areas. The model incorporating ETH input produces a visibly more accurate nowcast, exhibiting reduced intensity degradation, less spatial blurring over time, and good spatial alignment with the observed precipitation pattern in comparison to the model without ETH. In this case, the model was able to leverage the additional vertical structure information contained in the ETH to enhance nowcast quality.

Figure~\ref{fig:august} depicts a case featuring a prominent precipitation system located near the northeastern edge of the domain. The corresponding echo top height (ETH) field exhibits artifacts -- visible as circular bands -- resulting from the separate radar sweeps having different ranges and maximum observation altitudes. Despite these limitations, the model with ETH input appears to effectively utilize the available ETH information, maintaining the intensity of the precipitation signal over a longer lead time horizon. The nowcasts also appear to be less spatially blurry. This interesting case suggests a certain degree of robustness in the model's ability to extract useful features from imperfect ETH observations.

While limited to only three cases, these qualitative comparisons offer valuable insight into how echo top height can influence nowcast performance under different precipitation scenarios. In the May case, where ETH values were high and coincided with intense convective activity, the model with ETH produced more accurate and spatially coherent forecasts. In contrast, the February case - with high rainfall but low ETH values - only showed modest gains and a less accurate depiction of the main precipitation structure, showing that ETH may not always be informative or helpful, and sometimes even detrimental to accurately predicting the space-time structures and dynamics.

The August case, despite containing obvious artifacts in ETH estimates, still showed some benefit in including ETH, retaining the precipitation intensity and spatial sharpness of the field over longer lead times. It also showed a certain degree of robustness of the model predictions to imperfect and noisy ETH data.

Taken together, these qualitative evaluations suggest that while echo top height (ETH) input does not universally and systematically improve nowcasts, it can provide meaningful benefits under the right meteorological conditions. The three case studies suggest that ETH might be particularly helpful in scenarios with high ETH observations. While not explored in this work, splitting the test set and analyzing model performance across different ETH ranges may offer further insight and are worth considering in future studies.

\subsubsection*{Section Summary}
Eight model pairs with and without echo top height input were trained using varying random seeds and data splits.  The results show that overall, incorporating ETH leads to slightly larger output variance and underestimation bias. While the use of ETH tends to improve predictive performance during light precipitation events, especially at short lead times, it generally worsens performance in cases of widespread moderate rainfall and at higher precipitation intensities. Qualitative visual evaluations of selected case studies further support these findings, revealing large case-specific differences in predicted spatial structure and intensity between model versions. These examples suggest that ETH can provide valuable contextual information under certain conditions. Future work should therefore focus on identifying the scenarios and better understanding the conditions under which the addition of ETH is most likely to improve the predictions.

\section{Conclusion}
In this study, we explored the potential of incorporating radar-derived echo top height variable (ETH) into deep learning-based precipitation nowcasting. Using a deterministic U-Net model with 3D convolutions, we showed that ETH can be effectively integrated as an additional input channel.

However, while the models were able to incorporate echo top height (ETH) data, the results do not provide convincing proof-of-concept evidence of its value for precipitation nowcasting. Improvements were mostly limited to very low rain thresholds, and there were inconsistent and unpredictable effects on reducing issues such as rainfall intensity degradation and spatial blurring. Additionally, the inclusion of ETH introduced an additional negative bias in the forecasts, which is undesirable for applications such as flood forecasting and early warnings.

These findings suggest that although ETH carries some physically relevant information, its usefulness for improving the prediction of near-surface precipitation processes in the Netherlands appears to be limited. More sophisticated model designs, better ETH preprocessing, introduction of generative or probabilistic approaches may be needed to fully exploit the potential benefit while mitigating the downsides. As such, our study offers a useful starting point and diagnostic benchmark for future efforts in improving deep learning-based nowcasting with 3D radar-derived features.

In a broader sense, this work highlights the challenges of leveraging additional information in radar-based nowcasting. We believe this work serves as both a valuable template and a cautionary tale for the critical evaluation of not only echo top heights, but also other auxiliary variables, such as dual-polarization radar products or vertically integrated liquid, prior to their operational integration.

\begin{credits}
\subsubsection{Online Visualizations}
To better inspect the behavior and qualitative performance of the trained models, we refer the reader to a set of exported nowcast animations available at: \url{https://pavlikp.github.io/ETOPS-results/}.

\subsubsection{Code Repository}
All the training and evaluation code is available at \url{https://github.com/kinit-sk/EchoTops}.

\subsubsection{\ackname} This research was partially supported by TAILOR, a project funded by EU Horizon 2020 research and innovation programme under GA No 952215; by The Ministry of Education, Science, Research and Sport of the Slovak Republic under the Contract No. 0827/2021. 

Additionally, this research was partially supported by the Visegrad group for Vehicle to X (V4Grid), a project funded by Interreg Central Europe programme under CE0200803.

The authors acknowledge the use of computational resources of the DelftBlue supercomputer, provided by Delft High Performance Computing Centre (\url{https://www.tudelft.nl/dhpc})~\cite{DHPC2024}.

Research results were obtained using the computational resources procured in the national project National competence centre for high performance computing (project code: 311070AKF2) funded by European Regional Development Fund, EU Structural Funds Informatization of society, Operational Program Integrated Infrastructure.

\subsubsection{\discintname}
The authors have no competing interests to declare that are relevant to the content of this article.
\end{credits}
%
%
%
\bibliographystyle{splncs04}
\bibliography{references}

\end{document}